\documentclass[sigconf]{acmart}
\AtBeginDocument{%
  }

\setcopyright{acmlicensed}
\copyrightyear{2025}
\acmYear{2025}
\setcopyright{acmlicensed}\acmConference[MM '25]{Proceedings of the 33rd
ACM International Conference on Multimedia}{October 27--31, 2025}{Dublin,
Ireland}
\acmBooktitle{Proceedings of the 33rd ACM International Conference on
Multimedia (MM '25), October 27--31, 2025, Dublin, Ireland}
\acmDOI{10.1145/3746027.3755032}
\acmISBN{979-8-4007-2035-2/2025/10}

\usepackage{multirow}
\usepackage{booktabs}
\usepackage{graphicx} 
\usepackage{subcaption}
\usepackage{amsfonts}       
\usepackage{nicefrac}       
\usepackage{microtype}      
\usepackage{colortbl}
\usepackage{enumitem} 
\usepackage{caption}
\usepackage{multirow}
\usepackage{float}
\usepackage{pifont}
\usepackage{arydshln}
\usepackage{svg}
\usepackage{balance} 

\definecolor{color1}{RGB}{157,27,229}
\definecolor{color2}{RGB}{127,220,244} 
\definecolor{mygreen}{rgb}{0.1, 0.7, 0.2}
\definecolor{myred}{rgb}{139, 0, 0}
\definecolor{myorange}{rgb}{0.93, 0.51, 0.18}
\definecolor{LightCyan}{rgb}{0.96,0.96,0.96}
\definecolor{lightblue}{RGB}{240,240,240}

\begin{document}

\title{Decoupled Global-Local Alignment for Improving \\ Compositional Understanding}


\author{Xiaoxing Hu}
\authornote{Equal contribution.}
\email{xiaoxinghhh@gmail.com}
\orcid{1234-5678-9012}
\affiliation{%
  \institution{Beijing Institute of Technology}
  \department{School of Information and Electronics}
  \city{Beijing}
  \country{China}
}
\author{Kaicheng Yang}
\authornotemark[1] 
\email{kaichengyang@deepglint.com}
\affiliation{%
  \institution{DeepGlint} 
  \city{Beijing}
  \country{China}
}

\author{Jun Wang}
\email{junwang@deepglint.com}
\affiliation{%
  \institution{DeepGlint}
  \city{Beijing}
  \country{China}
}
\author{Haoran Xu}
\email{xhr964691257@163.com}
\affiliation{%
  \institution{Zhejiang University}
  \city{Zhejiang Province}
  \country{China}
}
\author{Ziyong Feng}
\email{ziyongfeng@deepglint.com}
\affiliation{%
  \institution{DeepGlint}
  \city{Beijing}
  \country{China}
}
\author{Yupei Wang}
\authornote{Corresponding author.}

\email{wangyupei2019@outlook.com}
\affiliation{%
  \institution{Beijing Institute of Technology}
 \department{School of Information and Electronics}
  \city{Beijing}
  \country{China}
}

\renewcommand{\shortauthors}{Xiaoxing Hu et al.}
\begin{abstract}
Contrastive Language-Image Pre-training~(CLIP) has achieved success on multiple downstream tasks by aligning image and text modalities. However, the nature of global contrastive learning limits CLIP's ability to comprehend compositional concepts, such as relations and attributes. Although recent studies employ global hard negative samples to improve compositional understanding, these methods significantly compromise the model's inherent general capabilities by forcibly distancing textual negative samples from images in the embedding space. To overcome this limitation, we introduce a \textbf{De}coupled \textbf{G}lobal-\textbf{L}ocal \textbf{A}lignment~(\textbf{DeGLA}) framework that improves compositional understanding while substantially mitigating losses in general capabilities. To optimize the retention of the model's inherent capabilities, we incorporate a self-distillation mechanism within the global alignment process, aligning the learnable image-text encoder with a frozen teacher model derived from an exponential moving average. Under the constraint of self-distillation, it effectively mitigates the catastrophic forgetting of pretrained knowledge during fine-tuning. To improve compositional understanding, we first leverage the in-context learning capability of Large Language Models (LLMs) to construct about 2M high-quality negative captions across five types. Subsequently, we propose the Image-Grounded Contrast (IGC) loss and Text-Grounded Contrast (TGC) loss to enhance vision-language compositionally. Experimental results across both general and compositional reasoning tasks validate the effectiveness of the DeGLA framework. Our code is released at \href{https://github.com/xiaoxing2001/DeGLA}{https://github.com/xiaoxing2001/DeGLA}.

\end{abstract}




\maketitle

\begin{figure}[htbp]
    \centering
    \begin{subfigure}[b]{0.49\columnwidth} 
        \centering
        \includegraphics[width=\textwidth]{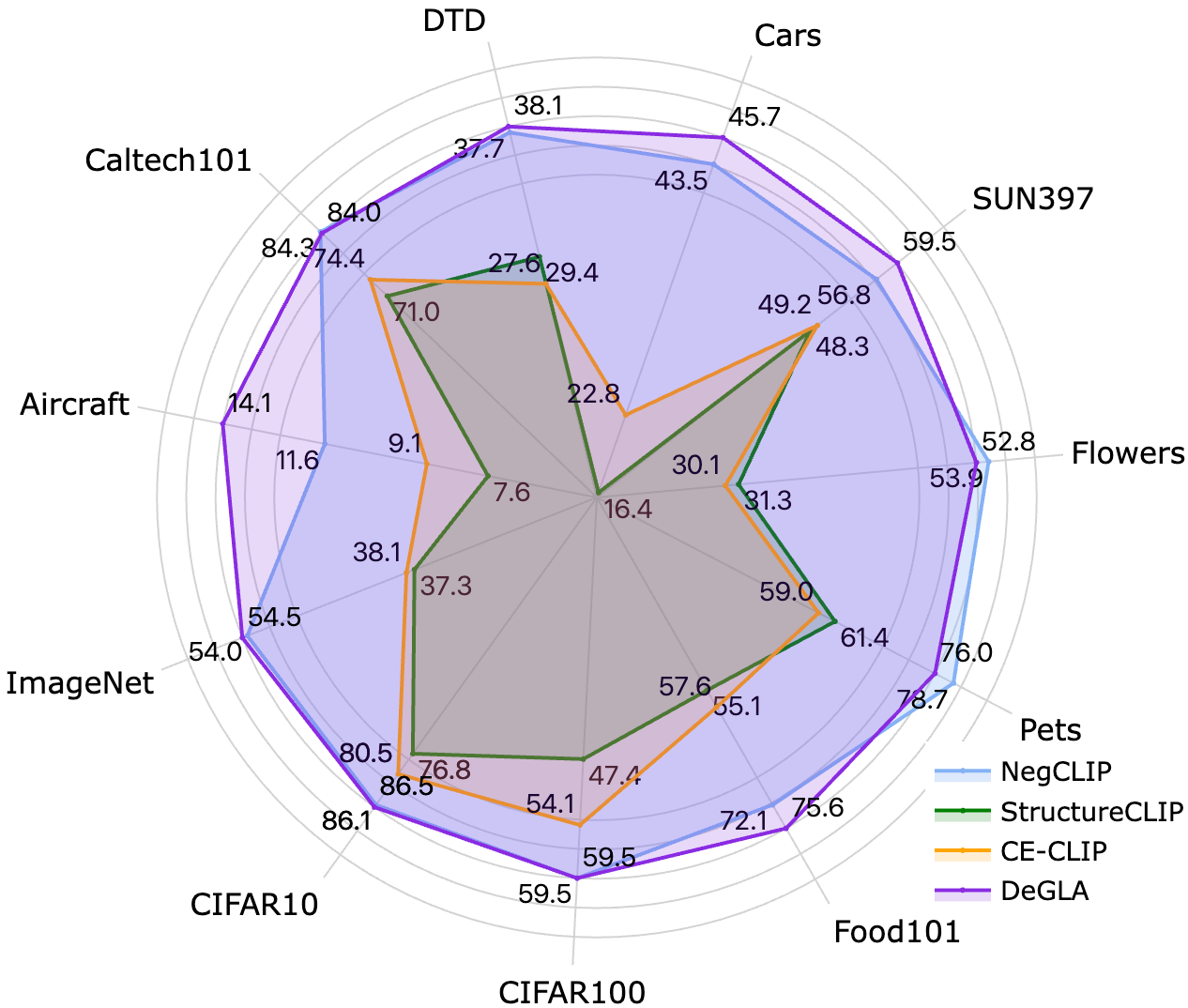} 
        \caption{General performance.}
        \label{fig:sub1}
    \end{subfigure}
    \begin{subfigure}[b]{0.5\columnwidth} 
        \centering
        \includegraphics[width=\textwidth]{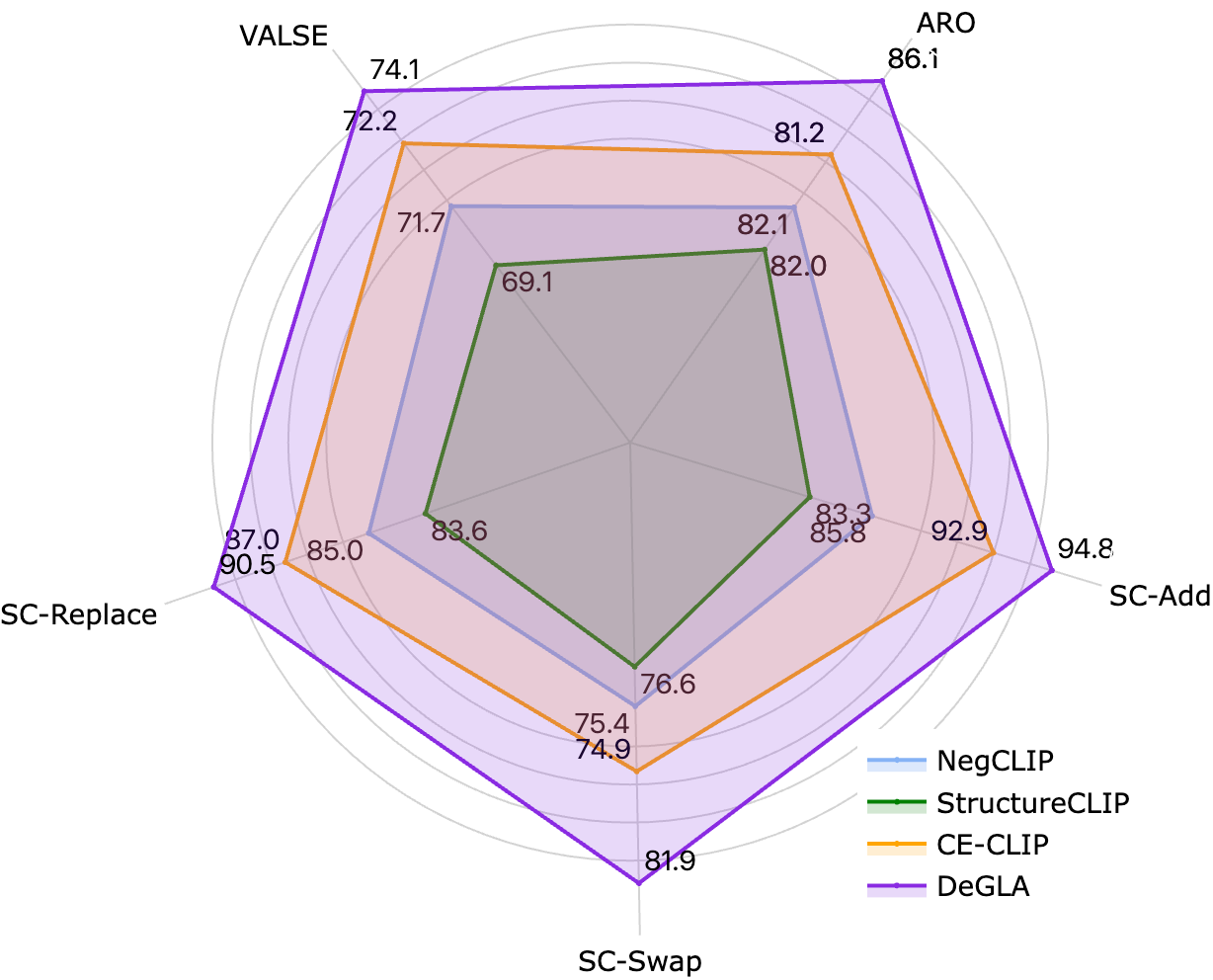} 
        \caption{Compositional performance}
        \label{fig:sub2}
    \end{subfigure}
    \begin{subfigure}[b]{\columnwidth} 
        \centering
        \includegraphics[width=1.0\columnwidth]{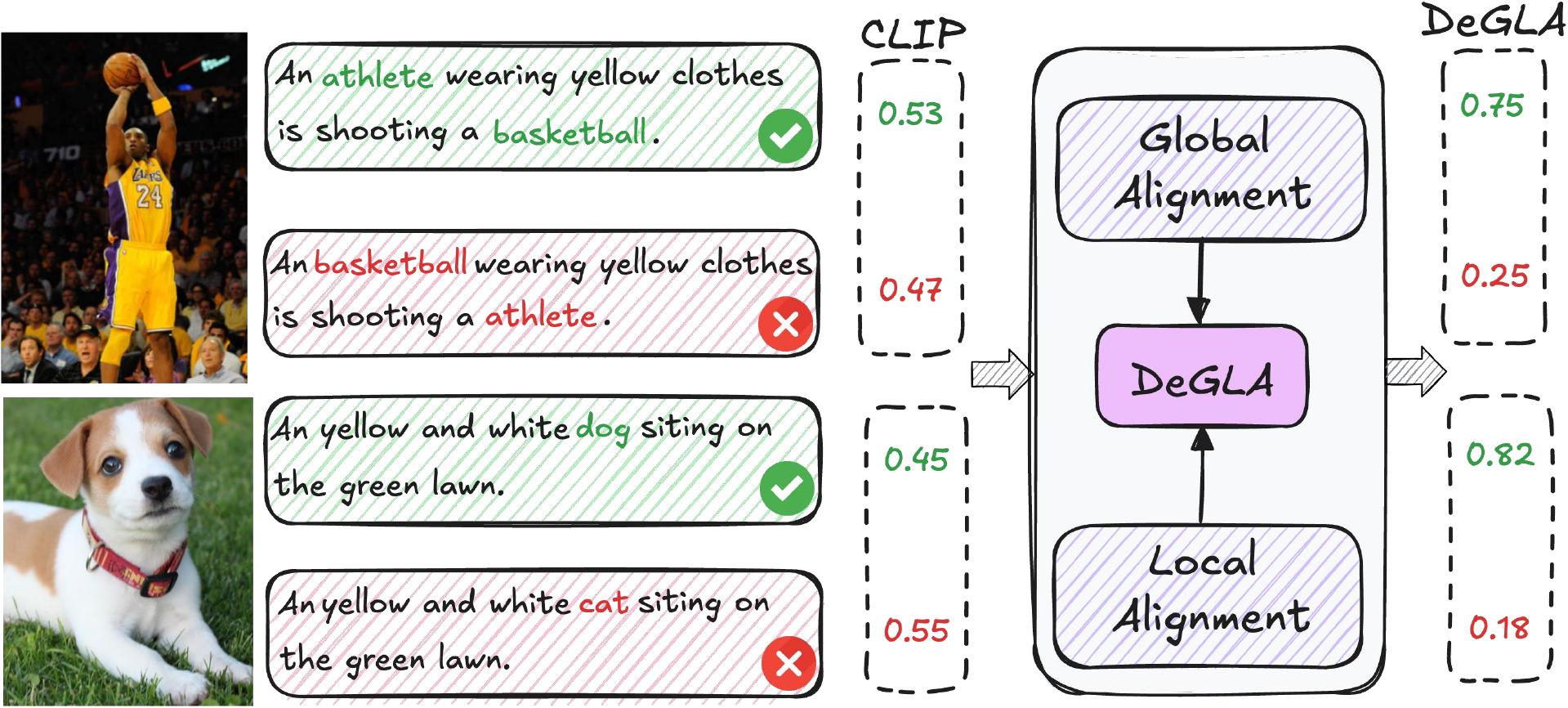} 
        \caption{An illustration for the compositional understanding.}
        \label{fig:sub3}
    \end{subfigure}
    \vspace{-5mm}
    \caption{(a) General performance comparison across 11 classification datasets. (b) Compositional performance comparison on VALSE, ARO, and SugarCrepe benchmarks. (c) CLIP scores for images with aligned and unaligned captions. DeGLA demonstrates significantly enhanced capabilities in compositional understanding.}
    \vspace{-3mm}
\end{figure}
\section{Introduction}
\label{sec:intro}

The rapid expansion of mobile networks and social platforms significantly boosts the large-scale generation of image-text pairs, laying a crucial foundation for vision-language pre-training~\cite{gu2025realsyn,gu2025breaking}. In particular, Contrastive Language-Image Pre-training~\cite{clip} (CLIP) leverages two distinct unimodal encoders for images and texts, and employs contrastive loss~\cite{infonce} for representation learning. After pretraining on extensive image-text pairs sourced from the internet, CLIP exhibits strong transferability and is extensively applied across various tasks, including image captioning~\cite{alayrac2022flamingo,li2023blip,manas2022mapl}, object detection~\cite{object_1,object_2,object_3}, and segmentation~\cite{li2022language,cho2024cat,yu2023convolutions,minderer2022simple,hoyer2025semivl,gu2021open,zang2022open,wu2024clip2uda}.

As CLIP attracts increasing attention from researchers, several enhanced methods based on CLIP have been proposed~\cite{flip,rwkvclip,zhang2024alignclip}. SLIP~\cite{mu2022slip} introduces a multitask learning framework that integrates self-supervised learning with CLIP pretraining. FILIP~\cite{yao2021filip} enhances the fine-grained alignment between image patches and textual words by refining the contrastive loss, while maintaining the capability for offline pre-computation of image and text representations during inference, ensuring efficiency in both large-scale training and inference phases. ALIP~\cite{yang2023alip} proposes the use of synthetic captions and adaptive contrastive loss to mitigate the influence of noisy data and improve the efficacy of pre-training data utilization. SigLIP~\cite{siglip} utilizes a sigmoid loss, which supports scaling up batch sizes and also performs effectively at smaller batch sizes. However, global contrastive learning inadequately leverages compositional structures within image-text pairs, thereby limiting CLIP's capability to capture nuanced compositional information in multimodal data~\cite{yuksekgonul2022and}. As shown in Figure~\ref{fig:sub3}, CLIP struggles to discern the relationship between ``athlete'' and ``basketball''.

Recent studies~\cite{yuksekgonul2022and,huang2024structure,zhang2024contrasting,doveh2023dense,doveh2023teaching} have sought to improve the compositional understanding of CLIP. NegCLIP~\cite{yuksekgonul2022and} introduces the Attribution, Relation, and Order (ARO) benchmark and, for the first time, proposes a fine-tuning framework that integrates hard negative samples to enhance CLIP's compositional understanding. Structure-CLIP~\cite{huang2024structure} integrates Scene Graph Knowledge (SGK) to augment multimodal structured representations. 
Hard-positives~\cite{hard_positive} incorporates hard positive samples during fine-tuning to strengthen the model’s capacity to capture subtle but semantically related variations among similar instances. CE-CLIP~\cite{zhang2024contrasting} introduces a simple yet effective fine-tuning framework with two fine-grained alignment losses to enhance compositional understanding, achieving state-of-the-art performance on multiple vision-language compositional reasoning benchmarks. \textit{However, our analysis reveals that existing methods improve CLIP's compositional understanding at the cost of general capabilities, exhibiting significant catastrophic forgetting of pre-trained knowledge}.

To address the limitation of previous works, we innovatively introduce a \textbf{De}coupled \textbf{G}lobal-\textbf{L}ocal \textbf{A}lignment (\textbf{DeGLA}) framework, which aims to improve compositional understanding while substantially mitigating losses in general capabilities. To optimize the retention of the inherent capabilities of the model, we incorporate a self-distillation mechanism within the global alignment process, aligning the learnable image-text encoder with a teacher model derived from an exponential moving average. To improve compositional understanding, we first leverage the in-context learning capability of Large Language Models (LLMs) to construct about 2M high-quality negative captions across five types. Subsequently, we propose the Image-Grounded Contrast (ICC) and Text-Grounded Contrast (TCC) to improve fine-grained understanding. We conduct extensive experiments and demonstrate that our method achieves new state-of-the-art performance in both compositional and general tasks. The main contributions are summarized as follows:

\begin{itemize}[leftmargin=*]
    \item We observe that previous methods, while enhancing CLIP's compositional understanding, often \textbf{compromise its general understanding capabilities}.
    \item We propose \textbf{a simple yet effective negative caption generation pipeline} that leverages the context learning capability of Large Language Models (LLMs) to generate high-quality negative captions.    
    \item We introduce the \textbf{DeGLA framework}, which employs a self-distillation mechanism within the global alignment to maintain the model's inherent general comprehension capabilities. Additionally, it combines Image-Grounded Contrast (IGC) loss and Text-Grounded Contrast (TGC) loss to improve vision-language compositional understanding.
    \item We conduct \textbf{extensive experiments and demonstrate that DeGLA achieves new state-of-the-art performance} in both compositional and general tasks.
 \end{itemize}

\section{Related work}
\label{sec:Related work}
\subsection{Vision-Language Contrastive Learning}

Vision-language representation learning aims to develop robust representations by pretraining models on extensive image-text pair datasets. A key development in this field, CLIP~\cite{radford2021learning}, pretrained on 400 million web-collected image-text pairs using contrastive learning, demonstrates strong zero-shot generalization across a wide range of vision tasks. Recent advancements include several methodologies that build on CLIP's framework. UniCLIP~\cite{lee2022uniclip} improves data efficiency by integrating contrastive losses from various domains into a single universal space. HiCLIP~\cite{geng2023hiclip} incorporates hierarchy-aware attention mechanisms in both visual and linguistic branches of CLIP, significantly enhancing cross-modal alignment. LaCLIP~\cite{fan2023improving} employs large language models to rewrite text, thereby increasing sentence structure and vocabulary diversity while preserving essential concepts and meanings. Nonetheless, these developments do not rectify the limitations in CLIP's compositional understanding, constrained by the inherent shortcomings of global contrastive learning~\cite{yuksekgonul2022and}.

\subsection{Vision-Language Compositionality}
Recently, there have been some works~\cite{yuksekgonul2022and,hsieh2024sugarcrepe,parcalabescu2021valse,thrush2022winoground,kamath2023s,peng2024synthesize} aim to enhance the compositional understanding of CLIP. NegCLIP~\cite{yuksekgonul2022and} introduces the Attribution, Relation, and Order (ARO) benchmark and utilizes hard negatives, consisting of nearest neighboring images within each batch, to force models to discern fine-grained differences in highly similar scenes. Structure-CLIP~\cite{huang2024structure} incorporates Scene Graph Knowledge (SGK) to enhance multimodal structured representations. CE-CLIP~\cite{zhang2024contrasting} proposes a simple yet effective strategy to optimize the utilization of existing image-text datasets, achieving state-of-the-art performance across multiple vision-language compositional reasoning benchmarks. However, although these methods significantly enhance CLIP's compositional understanding, they frequently compromise its original generalization capabilities. Developing an approach that simultaneously improves both general and compositional understanding continues to be a substantial challenge.

\subsection{Knowledge Distillation}
Knowledge distillation~\cite{hinton2015distilling} is extensively utilized in various domains, including vision~\cite{huang2024etag,li2021online,li2023curriculum} and natural language processing~\cite{jiao2019tinybert}. Recently, a series of knowledge distillation methods tailored specifically for CLIP have been proposed. TinyCLIP~\cite{wu2023tinyclip} introduces affinity mimicking, which explores the interaction between modalities during distillation, enabling student models to replicate the teacher's behavior in learning cross-modal feature alignment within a visual-linguistic affinity space. CLIP-KD~\cite{yang2024clip} proposes several distillation strategies, including relation, feature, gradient, and contrastive paradigms, aimed at maximizing feature similarity between the teacher and student models. CLIP-CID~\cite{yang2024clip} employs cluster-instance discrimination to facilitate knowledge transfer from the teacher model to the student model, thereby enabling the student model to develop a comprehensive semantic understanding of the pre-training data. Different from the above method, this paper introduces a self-distillation mechanism in global alignment to mitigate catastrophic forgetting during training and preserve the generalization capabilities of the model.

\begin{figure}[t!]
    \centering
    \includegraphics[width=1.0\linewidth]{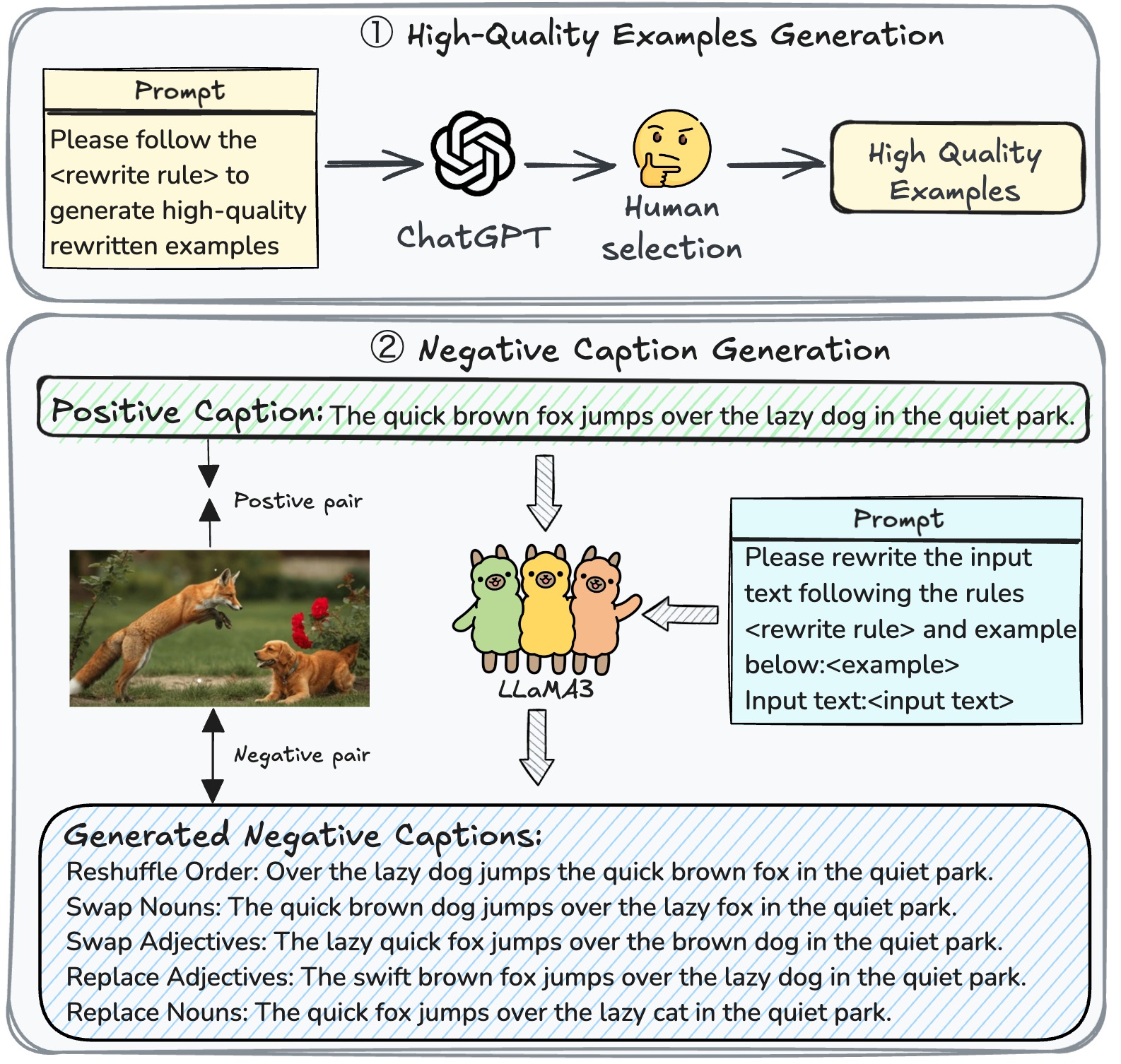}
    \vspace{-5mm}
    \caption{The overview of our proposed LLM-driven negative caption generation pipeline. We leverage the robust in-context learning capabilities of LLM to generate five types of hard negative captions.}
    \vspace{-5mm}
    \label{fig:negative_generation}
\end{figure}

\section{Method}

In subsequent sections, we first present the preliminary knowledge of CLIP in Section~\ref{subsec:preliminary}, followed by a detailed exposition of our proposed LLM-driven negative caption generation pipeline in Section~\ref{sec:negative data}. A comprehensive description of the DeGLA training framework is provided in Section~\ref{subsec:train framework}.

\subsection{Preliminaries of CLIP}\label{subsec:preliminary}
CLIP consists of a text encoder $\mathcal{E}_T$, and an image encoder $\mathcal{E}_I$. It encodes a batch of image-text pairs $\{ (I_i, T_i) \}_{i=1}^\mathcal{B}$ into the feature space $\{ (v_i, t_i) \}_{i=1}^\mathcal{B}$. The model employs the InfoNCE loss \cite{oord2018representation} to optimize the embeddings by minimizing the distance between corresponding image and text features while maximizing the distance between non-corresponding pairs. The image-to-text contrastive loss function is defined as follows:
\begin{equation}\label{Eq:image-to-text}
    \mathcal{L}_{I \rightarrow T} = -\log \frac{\exp(v_i \cdot t_i^\top / \tau)}{\sum_{j=1}^{\mathcal{B}} \exp(v_i \cdot t_j^\top / \tau)},
\end{equation}
where $v_i$ serves as the anchor. Given $t_i$ as the anchor, The symmetric text-to-image contrastive loss is formulated as follows:
\begin{equation}
    \mathcal{L}_{T\rightarrow I} = -\log \frac{\exp(t_i \cdot v_i^\top/\tau)}{\sum_{j=1}^{\mathcal{B}}\exp(t_i\cdot v_{j}^\top/\tau)},
\end{equation}
where $\cdot$ denotes the dot product used to calculate the similarity between normalized image and text embeddings, and $\tau$ is a learnable temperature parameter. The overall CLIP loss is:
\begin{equation}\label{eq:L_CLIP}
    \mathcal{L}_{CLIP} = \frac{1}{2}( \mathcal{L}_{I\rightarrow T}+\mathcal{L}_{T\rightarrow I}).
\end{equation}
CLIP encodes all semantic content from images and texts into a global representation during its training process, which constrains its capability for compositional understanding. While recent works~\cite{yuksekgonul2022and,huang2024structure,zhang2024contrasting} have sought to enhance this ability by introducing hard negative samples to promote the learning of local semantics, our analysis reveals experimental results demonstrate that such enhancements often degrade the model's inherent general comprehension capabilities. Consequently, this paper focuses on improving compositional understanding without compromising the original general comprehension abilities of the model.

\begin{table}[t!]
\caption{Detailed descriptions of the generated high-quality negative samples.}
\vspace{-1mm}
\label{tab:neg_text}
\centering
\resizebox{\columnwidth}{!}{ 
    \begin{tabular}{lll}
        \toprule
        \multicolumn{2}{c}{Negative types} & Rewrite rule \\
        \midrule
        \multirow{3}{*}{Intra-sentence reshuffle}
        
        & Subtype1 & Reshuffle the overall order of the sentence \\
        & Subtype2 & Swap the nouns in the sentence \\
        & Subtype3 & Swap the adjectives in the sentence \\
        \midrule
        \multirow{2}{*}{Minimal semantic substitution} & Subtype4 & Replace the adjectives in the sentence \\
        & Subtype5 & Replace the nouns in the sentence \\
        \bottomrule
    \end{tabular}
}
\vspace{-3mm}
\end{table}

\subsection{LLM-Driven Negative Caption Generation}\label{sec:negative data}
To improve compositional understanding of CLIP, we first leverage the in-context learning capability of Large Language Models (LLMs) to construct about 2M high-
quality negative captions across multiple types. 
Previous methods for generating compositional negative captions can be classified into two categories: rule-based generation~\cite{yuksekgonul2022and,huang2024structure} and unmasking-based generation~\cite{zhang2024contrasting,doveh2023teaching,doveh2023dense}. However, rule-based approaches are limited in their ability to produce complex compositional negatives, such as those involving minimal semantic substitutions. In contrast, unmasking-based methods can generate diverse negatives but often fail to avoid producing hard positive samples such as ``the cat is eating the food'' $\rightarrow$ ``the cat is eating the hot food''. 

To address these limitations, we propose an LLM-driven generation method to enhance and standardize the generation of compositional negative captions.  Specifically, we categorize negative captions into two primary types: (1) intra-sentence reshuffling, which improves the VLM's sensitivity to sentence order, and (2) minimal semantic substitution, which enhances the VLM's ability to discern subtle semantic variations. For intra-sentence reshuffling, we examine three subtypes: (a) full sentence reordering, (b) noun swapping, and (c) adjective swapping. For minimal semantic substitution, we investigate two subtypes: (a) adjective replacement and (b) noun replacement. Detailed descriptions of each negative sample type are presented in Table~\ref{tab:neg_text}. As shown in Figure~\ref{fig:negative_generation}, we initially generate high-quality negative samples to fully leverage the in-context learning capabilities of LLMs. For each type, we first use ChatGPT4-Turbo to generate 200 rewritten examples according to rewritten rules, then manually select the 50 highest-quality examples through rigorous evaluation. Building upon these examples, we employ the Llama-3.1-8B-Instruct model\footnote{\url{https://huggingface.co/meta-llama/Llama-3.1-8B-Instruct}} to generate large-scale compositional negatives. To ensure semantic divergence from original sentences and avoid hard positives, we incorporate explicit filtering constraints in the prompts (e.g., ``generated sentences must exhibit distinct semantics''). Complete prompt templates and representative samples are provided in the appendix.  


\subsection{DeGLA Framework}
\label{subsec:train framework}

\textbf{Global Alignment.}
Building upon the base setting~\cite{yuksekgonul2022and}, we integrate negative captions into the Equation ~\ref{Eq:image-to-text} for global alignment, yielding the augmented image-to-text contrastive:
\begin{equation}\label{eq:aug_image-to-text}
    \mathcal{L}_{I \rightarrow T} = -  \log \frac{\exp(v_i \cdot t_i^\top / \tau)}{ \sum_{j=1}^\mathcal{B} \left( \exp(v_i \cdot t_j^\top / \tau)+\sum_{k=1}^\mathcal{K}\exp(v_i\cdot \bar{t}_{i,k}^\top/\tau) \right) } ,
\end{equation}
where $\bar{t}_{j,k}$ denotes the $k$-th negative text related to the $j$-th text. In this work, we set $\mathcal{K}=4$ to denote the number of negative samples paired with each positive sample during training, due to the merging of two subtypes, as detailed in the appendix. Then we utilize the hard negative-aware image-text contrastive loss as our base loss which can be formulated as:
\begin{equation}
    \mathcal{L}_{base} = \frac{1}{2}( \mathcal{L}_{I\rightarrow T}+\mathcal{L}_{T\rightarrow I}),
\end{equation}
However, directly incorporating hard negative text into global contrastive learning significantly compromises the model's general capabilities, as indicated in Section~\ref{sec:exp}. In this paper, we introduce a self-distillation mechanism within the global alignment framework to mitigate the catastrophic forgetting of pre-trained knowledge. To enhance model robustness and prevent rapid forgetting of pre-trained knowledge during training, we employ an Exponential Moving Average (EMA) strategy to update the weights of the frozen image encoder $\mathcal{E}_I^*$ and text encoder $\mathcal{E}_T^*$:
\begin{equation}\label{eq:ema}
\begin{aligned}
    \mathcal{E}_I^* &= \alpha \mathcal{E}_I^*+(1-\alpha)\mathcal{E}_I , \\
    \mathcal{E}_T^* &= \alpha \mathcal{E}_T^*+(1-\alpha)\mathcal{E}_T ,
\end{aligned}
\end{equation}
where the hyperparameter $\alpha$ controls the update speed of the frozen model's parameters. Given a batch of image-text pairs $\{I_i, T_i,\bar{T}_{i}\}$, we obtain the image and text embeddings $\{v_i, t_i,\bar{t}_{i}\}$ from the learnable encoders ($\mathcal{E}_I, \mathcal{E}_T$) and $\{v_i^*, t_i^*,\bar{t}_{i}^{*}\}$ from the frozen EMA encoders ($\mathcal{E}_I^*, \mathcal{E}_T^*$), respectively. Following L2 normalization, we compute the squared distance between the image-text embeddings from the learnable and frozen EMA encoders to quantify the knowledge discrepancy between the fine-tuned and pre-trained models:
\begin{equation}\label{eq:distill}
    \mathcal{L}_{Distill} = \sum_{i=1}^\mathcal{B} \left( \left\| v_i - v_i^* \right\|_2^2 + \left\| t_i - t_i^* \right\|_2^2
    + \sum_{k=1}^\mathcal{K} \left\| \bar{t}_{i,k} - \bar{t}_{i,k}^* \right\|_2^2 \right).
\end{equation}
Under the constraint of Equation~\ref{eq:distill}, the image-text embeddings undergo subtle refinements within the pre-trained representation space, thereby preserving the model's general capabilities from excessive degradation.

\begin{figure}[t!]
    \centering
    \includegraphics[width=\linewidth]{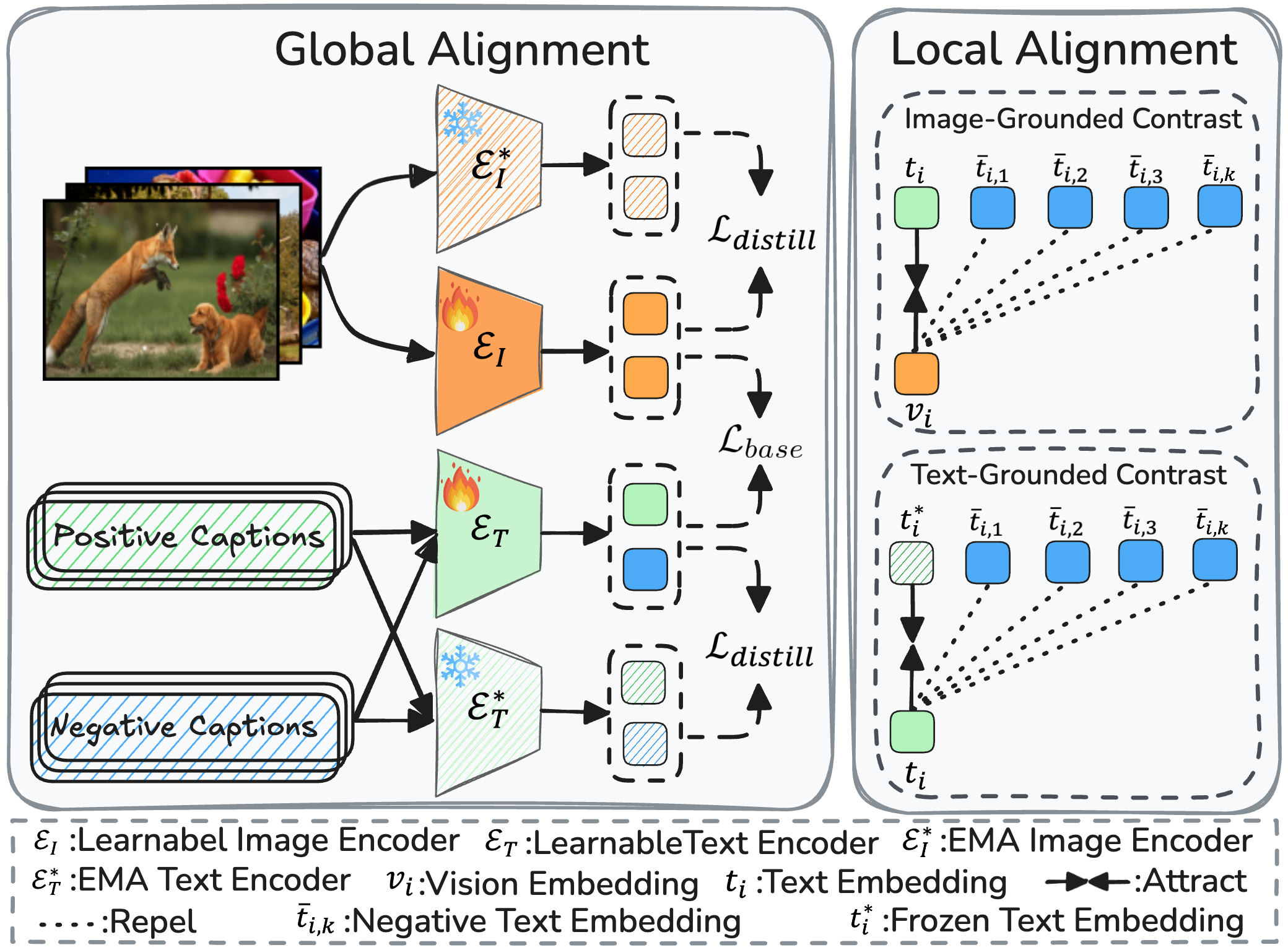}
    \vspace{-5mm}
    \caption{The proposed training framework of DeGLA.}
    \label{fig:overview}
    \vspace{-3mm}
\end{figure}


\begin{table*}[t!]
\centering 
\caption{Results (\%) on VALSE. The best results are marked in bold,
and the second-best results are underlined. The \textcolor{mygreen}{number} represents the improvement of our method compared to the CE-CLIP.} 
\vspace{-3mm}
\label{tab:valse}
\resizebox{\textwidth}{!}{
\begin{tabular}{lccccccccccc}
\toprule
\multirow{2}{*}{\bf Model} & \multirow{2}{*}{\bf \#Params} &  \bf Existence &\bf Plurality   &\bf Counting &\bf Sp.rel.& \multicolumn{2}{c}{\bf Actions}&\multicolumn{2}{c}{\bf Coreference}&\multirow{2}{*}{\bf  Foil-it!} &\multirow{2}{*}{\bf  Avg.}\\
& &  quantifiers & number & & relations & repl. & actant swap & standard & clean & \\
\midrule

\textcolor{gray}{BLIP\cite{li2022blip}} &\textcolor{gray}{583M} & \textcolor{gray}{86.3}&\textcolor{gray}{73.2}&\textcolor{gray}{68.1}&\textcolor{gray}{71.5}&\textcolor{gray}{77.2}&\textcolor{gray}{61.1}&\textcolor{gray}{53.8}&\textcolor{gray}{48.2}&\textcolor{gray}{93.8}&\textcolor{gray}{70.0} \\
\textcolor{gray}{BEIT3\cite{wang2022image}} &\textcolor{gray}{1.9B}&\textcolor{gray}{77.4} & \textcolor{gray}{74.6}& \textcolor{gray}{68.8}&\textcolor{gray}{74.0}&\textcolor{gray}{86.7}&\textcolor{gray}{65.2}&\textcolor{gray}{50.0}&\textcolor{gray}{44.2}&\textcolor{gray}{96.0}&\textcolor{gray}{70.4} \\
\textcolor{gray}{BLIP2\cite{li2023blip}} &\textcolor{gray}{3.4B}&\textcolor{gray}{55.5}&\textcolor{gray}{71.5}&\textcolor{gray}{66.0}&\textcolor{gray}{62.4}&\textcolor{gray}{83.6}&\textcolor{gray}{51.6}&\textcolor{gray}{48.6}&\textcolor{gray}{51.9}&\textcolor{gray}{95.9} & \textcolor{gray}{65.4} \\
\textcolor{gray}{MiniGPT-4\cite{zhu2023minigpt}}&\textcolor{gray}{>9B}&\textcolor{gray}{65.5}&\textcolor{gray}{72.5}&\textcolor{gray}{67.4}&\textcolor{gray}{68.4}&\textcolor{gray}{83.2}&\textcolor{gray}{58.8}&\textcolor{gray}{52.6}&\textcolor{gray}{51.0}&\textcolor{gray}{95.8} & \textcolor{gray}{68.4} \\
\midrule
\multicolumn{4}{l}{\textit{Hard Negative based method}} & & & &  & & &\\

XVLM-coco\cite{xvlm}&216M&83.0&75.6&67.5&70.2&73.8&68.6&46.4&49.6&94.8&69.5 \\
CE-XVLM\cite{zhang2024contrasting}&216M&83.5&72.8&72.1&68.7&71.8&69.1&51.0&46.8&93.8 &70.8\\
\hdashline
CLIP\cite{radford2021learning}&151M&68.7&57.1&61.0&65.4&77.8&71.8&54.1&51.0&89.8&65.3\\
CyCLIP\cite{goel2022cyclip} &151M& 69.3 & 58.3&61.0&66.4&78.1&72.0&53.2&51.6&88.8&65.5 \\
NegCLIP\cite{yuksekgonul2022and}&151M& 76.8&72.0&\underline{65.2}&72.7&\underline{81.6}&84.8& \textbf{58.9}&\underline{54.8}&91.8&71.7\\
Structure-CLIP~\cite{huang2024structure}&151M&75.6 & 67.1 & 62.0 & 68.2 & 80.4 & 88.3 & 44.5 & \textbf{58.7} & 91.2 & 69.1\\
CE-CLIP\cite{zhang2024contrasting}&151M& \underline{78.6}& \textbf{77.6}&64.3& \underline{74.0}&81.2& \underline{88.5}&54.9& 52.9& \underline{93.7}&\underline{72.2}\\
\rowcolor{gray!20}
DeGLA~(ours) &151M& \textbf{82.4} & \underline{73.8} & \textbf{68.3} & \textbf{75.3} & \textbf{82.6} & \textbf{88.8} & \underline{58.5} & \underline{54.8} & \textbf{93.8} & \textbf{74.1(\textcolor{mygreen}{+1.9})}\\

\bottomrule
\end{tabular}}
\vspace{-3mm}
\end{table*}
\begin{table*}[t!]
\centering
\caption{Results(\%) on SugarCrepe. Vera and Grammar are text-only models.  The best results are marked in bold, and the second-best results are underlined. The \textcolor{mygreen}{number} represents the improvement of our method compared to the CE-CLIP.}
\vspace{-3mm}
\label{tab:sugarcrepe}
\resizebox{0.9\linewidth}{!}{
\begin{tabular}{lcccccccccc}
\toprule
 \multirow{2}{*}{\bf Model} &\multicolumn{4}{c}{\bf REPLACE} &\multicolumn{3}{c}{\bf SWAP }&\multicolumn{3}{c}{\bf ADD } \\
 \cmidrule(rrrr){2-5}  \cmidrule(rrr){6-8}\cmidrule(rrr){9-11}
 & Object&Attribute&Relation& Avg. & Object&Attribute& Avg.& Object&Attribute& Avg. \\
\midrule
\textcolor{gray}{Human} & \textcolor{gray}{100.0} & \textcolor{gray}{99.0} & \textcolor{gray}{97.0} & \textcolor{gray}{98.7} & \textcolor{gray}{99.0} & \textcolor{gray}{100.0} & \textcolor{gray}{99.5} & \textcolor{gray}{99.0} & \textcolor{gray}{99.0} & \textcolor{gray}{99.0} \\
\midrule

\textcolor{gray}{Vera~\cite{vera}} & \textcolor{gray}{49.4} & \textcolor{gray}{49.6} & \textcolor{gray}{49.1} & \textcolor{gray}{49.4} & \textcolor{gray}{49.4} & \textcolor{gray}{49.2} & \textcolor{gray}{49.3} & \textcolor{gray}{49.4} & \textcolor{gray}{49.6} & \textcolor{gray}{49.5} \\
\textcolor{gray}{Grammar~\cite{Grammar}} & \textcolor{gray}{50.0} & \textcolor{gray}{50.0} & \textcolor{gray}{50.0} & \textcolor{gray}{50.0} & \textcolor{gray}{50.0} & \textcolor{gray}{50.0} & \textcolor{gray}{50.0} & \textcolor{gray}{50.0} & \textcolor{gray}{50.0} & \textcolor{gray}{50.0} \\
\textcolor{gray}{BLIP2~\cite{li2023blip}} & \textcolor{gray}{-} & \textcolor{gray}{-} & \textcolor{gray}{-} & \textcolor{gray}{86.7} & \textcolor{gray}{-} & \textcolor{gray}{-} & \textcolor{gray}{69.8} & \textcolor{gray}{-} & \textcolor{gray}{-} & \textcolor{gray}{86.5} \\
\midrule
\multicolumn{4}{l}{\textit{Hard Negative based method}} & & & &  & & & \\
CLIP~\cite{radford2021learning} & 90.9 & 80.0 & 69.2 & 80.2 & 62.7 & 61.4 & 64.0 & 77.2 & 68.2 & 72.7 \\
NegCLIP~\cite{yuksekgonul2022and} & 92.7 & 85.9 & 76.5 & 85.0 & 75.3 & 75.2 & 75.4 & 88.8 & 82.8 & 85.8 \\
Structure-CLIP~\cite{huang2024structure} & 91.4 & 85.0 & 74.4 & 83.6 & 72.7 & \underline{80.5} & \underline{76.6} & 85.5 & 81.1 & 83.3\\
CE-CLIP~\cite{zhang2024contrasting} & \underline{93.1} & \underline{88.8} & \underline{79.0} & \underline{87.0} & \underline{72.8} & 77.0 & 74.9 & 92.4 & \underline{93.4} & \underline{92.9} \\
\rowcolor{gray!20}
DeGLA~(ours) & \textbf{94.5} & \textbf{92.6} & \textbf{84.2} & \textbf{90.5(\textcolor{mygreen}{+3.5})} & \textbf{81.6} & \textbf{82.1} & \textbf{81.9(\textcolor{mygreen}{+6.9})} & \textbf{93.8} & \textbf{95.7} & \textbf{94.8(\textcolor{mygreen}{+1.9})}\\ 
\bottomrule
\end{tabular}}
\vspace{-3mm}
\end{table*}
\begin{table}[t!]
\centering 
\caption{\textbf{Results (\%) on ARO}. The best results are marked in bold,
and the second-best results are underlined. The \textcolor{mygreen}{number} represents the improvement of our method compared to the CE-CLIP.}
\vspace{-3mm}
\label{tab:aro}
\resizebox{0.48\textwidth}{!}{
\begin{tabular}{lccccc}
\toprule
\bf Model & \textbf{Relation} & \textbf{Attribute}& \textbf{COCO-order}& \textbf{Flickr-order} & \textbf{Avg.} \\
\midrule

\textcolor{gray}{BLIP\cite{li2022blip}} &\textcolor{gray}{59.0}&\textcolor{gray}{88.0}&\textcolor{gray}{-}&\textcolor{gray}{-}&\textcolor{gray}{-} \\
\textcolor{gray}{BEIT3\cite{wang2022image}} &\textcolor{gray}{60.6}&\textcolor{gray}{74.6}&\textcolor{gray}{-}&\textcolor{gray}{-}&\textcolor{gray}{-} \\
\textcolor{gray}{BLIP2\cite{li2023blip}}&\textcolor{gray}{41.2}&\textcolor{gray}{71.3}&\textcolor{gray}{-}&\textcolor{gray}{-}&\textcolor{gray}{-} \\
\textcolor{gray}{MiniGPT-4\cite{zhu2023minigpt}} &\textcolor{gray}{46.9}&\textcolor{gray}{55.7}&\textcolor{gray}{-}&\textcolor{gray}{-}&\textcolor{gray}{-} \\
\midrule
\multicolumn{6}{l}{\textit{Hard Negative based method}}  \\
CLIP~\cite{radford2021learning} & 59.2 & 62.9 & 48.4 & 59.1 & 57.4 \\
CyCLIP\cite{goel2022cyclip}& 59.1 &65.4&-&-&- \\
NegCLIP\cite{yuksekgonul2022and}&80.4&70.5&\underline{86.9}&\underline{90.5}&\underline{82.1}\\
Structure-CLIP~\cite{huang2024structure}&\underline{81.8}&\textbf{80.5}&81.7&83.9&82.0 \\
CE-CLIP\cite{zhang2024contrasting}&\textbf{83.9}&\underline{76.4}&80.9&83.7&81.2\\
\rowcolor{gray!20}
DeGLA~(ours) &81.6& 74.3 & \textbf{93.8} & \textbf{94.7} & \textbf{86.1(\textcolor{mygreen}{+4.9})} \\
\bottomrule

\end{tabular}}
\vspace{-5mm}

\end{table}

\noindent\textbf{Local Alignment.} 
In addition to the global alignment of $\mathcal{L}_{\text{base}}$, to further enhance compositional understanding, we propose a Local Alignment, which includes Image-Grounded Contrast Loss(IGC) and Text-Grounded Contrast Loss(TGC).
We initially introduce the Image-Grounded Contrast loss to attract image embeddings towards positive text embeddings and repel them from negative text embeddings in the feature space~(as shown in Figure~\ref{fig:overview}). Specifically, given an image-text pair $(I_i, T_i)$ and corresponding hard negative texts $\bar{T}_{i}$, the IGC loss $\mathcal{L}_{IGC}$ is defined as follows:
\begin{equation}\label{eq:L_ICC}
    \mathcal{L}_{IGC} = -\log \frac{\exp(v_i\cdot t_i^\top/\tau)}{\exp(v_i\cdot t_i^\top/\tau)+\sum_{k=1}^\mathcal{K}\exp(v_i\cdot \bar{t}_{i,k}^\top/\tau)}),
\end{equation}
where $\tau$ is a temperature parameter that controls the sharpness of the distribution. Switching to the textual perspective, hard negative texts, which are minor modifications of the originals, remain proximate to the positive texts within the feature space of the CLIP text encoder. This closeness may lead to mismatches during image embedding alignment with positive text embeddings, potentially impairing compositional understanding. Different from the Image-Grounded Contrast loss, the Text-Grounded Contrast (TGC) loss $\mathcal{L}_{TGC}$ operates solely within the text modality. Its function is to enhance the text encoder's ability to more effectively discriminate between positive and negative texts, thus improving the compositional understanding of vision language models. The $\mathcal{L}_{TGC}$ is formulated as:
\begin{equation}\label{eq:L_TGC}
    \mathcal{L}_{TGC} = -\log \frac{\exp(t_i\cdot t_i^{*\top}/\tau)}{\exp(t_i\cdot t_i^{*\top}/\tau)+\sum_{k=1}^\mathcal{K}\exp(t_i\cdot \bar{t}_{i,k}^\top/\tau)},
\end{equation}
where $t_i^*$ represents the progressively frozen text embedding obtained from the frozen EMA text encoder $\mathcal{E}_T^*$. This embedding is used as an anchor in local alignment for two main reasons:  (1) It acts as the positive sample in contrastive learning, enabling the model to discriminate between positive and compositional negative samples; and (2) It mitigates potential overfitting of the text encoder to the feature space of the fine-tuning dataset. Finally, the overall loss function is defined as:
\begin{equation}
\label{eq:loss_final}
    \mathcal{L}_{all} = \mathcal{L}_{Base} + \lambda_1 \mathcal{L}_{IGC} + \lambda_2 \mathcal{L}_{TGC} + \lambda_3 \mathcal{L}_{Distill} .
\end{equation}
where $\lambda_1, \lambda_2, \lambda_3$ are loss weights to balance the influence of different loss functions.

\section{Experiments} 
\label{sec:exp}

\begin{table*}[t!]
\centering
\caption{Zero-shot classification performance on 11 datasets. The best results are marked in bold, and the second-best results are underlined. The \textcolor{mygreen}{number} represents the improvement of our method compared to the CE-CLIP.}
\vspace{-3mm}
\label{tab:classification}
\resizebox{\textwidth}{!}{
\begin{tabular}{lcccccccccccc}
\toprule
\textbf{Model}  & \textbf{CIFAR10} & \textbf{CIFAR100} & \textbf{Food101} & \textbf{Pets} & \textbf{Flowers} & \textbf{SUN397} & \textbf{Cars} & \textbf{DTD} & \textbf{Caltech101} & \textbf{Aircraft}  & \textbf{ImageNet} & \textbf{Avg.} \\
\midrule
\multicolumn{2}{l}{\textit{Pretrained model}} & & & &  & & &\\
\textcolor{gray}{CLIP~\cite{radford2021learning}}  & \textcolor{gray}{86.5} & \textcolor{gray}{61.0} & \textcolor{gray}{78.5} & \textcolor{gray}{79.6} & \textcolor{gray}{58.4} & \textcolor{gray}{59.9} & \textcolor{gray}{48.8} & \textcolor{gray}{38.7} & \textcolor{gray}{86.3} & \textcolor{gray}{15.3} & \textcolor{gray}{57.9} & \textcolor{gray}{61.0}\\ 
\midrule
\multicolumn{4}{l}{\textit{Hard negative based method}} & & & &  & & &\\
NegCLIP~\cite{yuksekgonul2022and}
 & \underline{86.1} & \textbf{59.9} & \underline{72.1} & \textbf{78.7} & \textbf{53.9} & \underline{56.8} & \underline{43.5} & \underline{37.7} & \textbf{84.3} & \underline{11.6} & \underline{54.0}& \underline{58.1}\\
 Structure-CLIP~\cite{huang2024structure} & 76.8 & 47.4 & 55.1 & 61.4 & 31.3 & 48.3 & 16.4 & 29.4 & 71.0 & 7.6 & 37.3 & 43.8\\
CE-CLIP~\cite{zhang2024contrasting} & 80.5 & 54.1  & 57.6 & 59.0 & 30.1 & 49.2 & 22.8 & 27.6 & 74.4 &9.1 & 38.1& 45.7\\
\rowcolor{gray!20}
DeGLA~(ours) & \textbf{86.5} & \underline{59.5} & \textbf{75.6} & \underline{76.0} & \underline{52.8} & \textbf{59.5} & \textbf{45.7} & \textbf{38.1}& \underline{84.0} & \textbf{14.1} & \textbf{54.5}& \textbf{58.7(\textcolor{mygreen}{+13.0})} \\

\bottomrule
\end{tabular}
}
\vspace{-3mm}
\end{table*}
\begin{table*}[t!]
\centering
\caption{Liner probe performance on 11 datasets. The best results are marked in bold, and the second-best results are underlined. The \textcolor{mygreen}{number} represents the improvement of our method compared to the CE-CLIP.}
\vspace{-3mm}
\label{tab:liner}
\resizebox{\textwidth}{!}{
\begin{tabular}{lcccccccccccc}
\toprule
\textbf{Model}  & \textbf{CIFAR10} & \textbf{CIFAR100} & \textbf{Food101} & \textbf{Pets} & \textbf{Flowers} & \textbf{SUN397} & \textbf{Cars} & \textbf{DTD} & \textbf{Caltech101} & \textbf{Aircraft}  & \textbf{ImageNet} & \textbf{Avg.} \\
\midrule
\multicolumn{2}{l}{\textit{Pretrained model}} & & & &  & & &\\
\textcolor{gray}{CLIP~\cite{radford2021learning}}  & \textcolor{gray}{95.0} & \textcolor{gray}{80.1} & \textcolor{gray}{88.5} & \textcolor{gray}{89.3} & \textcolor{gray}{94.6} & \textcolor{gray}{74.1} & \textcolor{gray}{80.8} & \textcolor{gray}{73.6} & \textcolor{gray}{90.5} & \textcolor{gray}{44.8} & \textcolor{gray}{74.3} & \textcolor{gray}{80.5}\\ 
\midrule
\multicolumn{4}{l}{\textit{Hard negative based method}} & & & &  & & &\\
NegCLIP~\cite{yuksekgonul2022and}
 & \underline{94.6} & \underline{80.0} & \underline{86.1} & \textbf{89.6} & \underline{93.9} & \underline{72.9} & \textbf{78.8} & \underline{72.9} & \textbf{90.0} & \underline{43.2} & \underline{72.9} & \underline{79.5}\\
 Structure-CLIP~\cite{huang2024structure} & 91.9 & 75.5 & 81.2 & 86.2 & 89.6 & 69.0 & 67.4 & 67.7 & 65.2 & 37.7 & 67.7 & 72.7\\
CE-CLIP~\cite{zhang2024contrasting} & 94.3 & 78.5  & 84.3 & 88.1 & 92.6 & 71.0 & \underline{74.1} & 71.8 & 88.3&39.6 & 70.7& 77.6\\
\rowcolor{gray!20}
DeGLA~(ours) & \textbf{95.1} & \textbf{80.5} & \textbf{86.7} & \underline{89.5} & \textbf{94.6} & \textbf{74.0} & \textbf{78.8} & \textbf{73.0}& \underline{89.6} & \textbf{43.5} & \textbf{73.4}& \textbf{79.9(\textcolor{mygreen}{+2.3})} \\

\bottomrule
\end{tabular}
}
\vspace{-3mm}
\end{table*}

\subsection{Implement details}
\label{subsec:implement_detail}

\noindent\textbf{Training Setup.} To ensure a fair comparison with previous studies~\cite{yuksekgonul2022and,zhang2024contrasting}, we employ LLaMA3.1-instruct-8B~\cite{llama3.1} to generate compositional negative samples from the MSCOCO dataset, facilitating direct comparisons with NegCLIP~\cite{yuksekgonul2022and} and CE-CLIP~\cite{zhang2024contrasting}. We utilize the CLIP-ViT/B-32 model as the foundation vision-language model, initializing it with pretrained weights from CLIP~\cite{radford2021learning}. The model is fine-tuned on 8 NVIDIA V100 (32G) GPUs for 5 epochs using a batch size of 256, consistent with the protocols of previous works~\cite{zhang2024contrasting, yuksekgonul2022and, doveh2023teaching}. We employ AdamW as the optimizer, initialized with a learning rate of $1 \times 10^{-6}$ and a weight decay of 0.1. The parameters $\alpha$, $\beta_1$, and $\beta_2$ are set to 0.9996, 0.9, and 0.98, respectively. We perform a hyperparameter search for $\lambda_1, \lambda_2, \lambda_3$, with optimal values of $\lambda_1 = 0.1$, $\lambda_2 = 0.1$, and $\lambda_3 = 0.005$.

\begin{figure}[t!]
    \centering
    \begin{subfigure}[b]{0.495\columnwidth} 
        \centering
        \includegraphics[width=\textwidth]{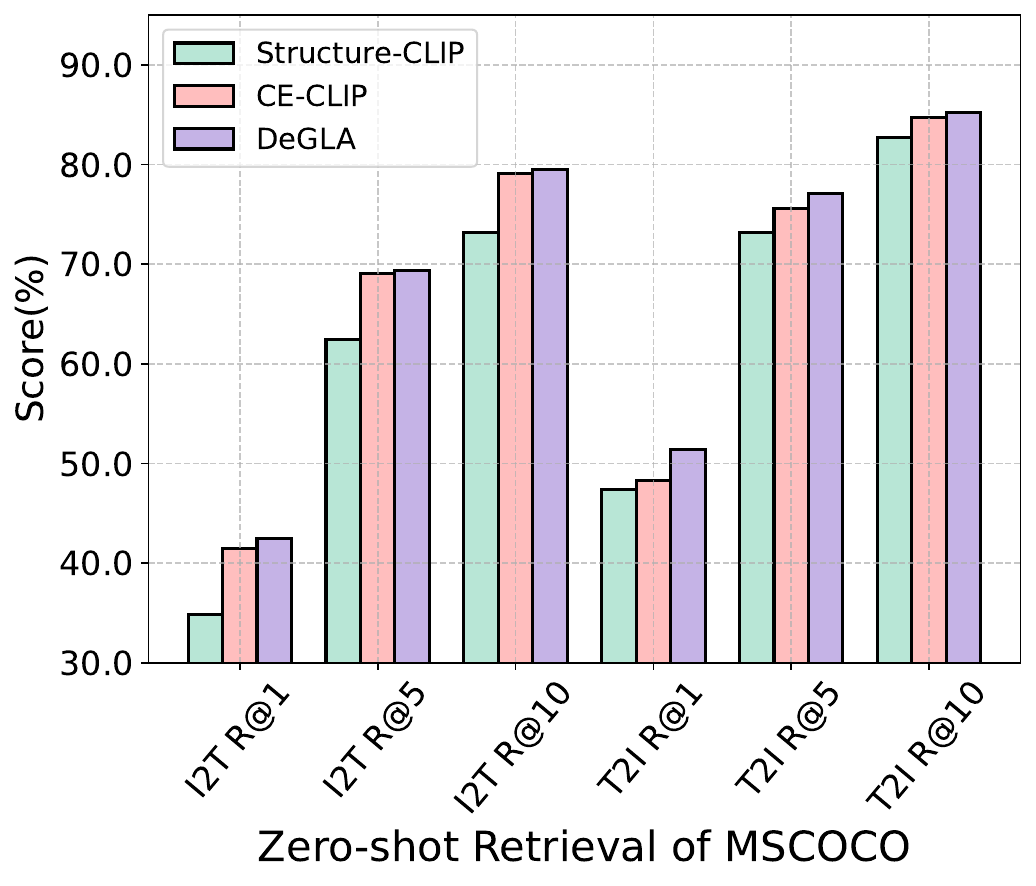} 
    \end{subfigure}
    \begin{subfigure}[b]{0.495\columnwidth} 
        \centering
        \includegraphics[width=\textwidth]{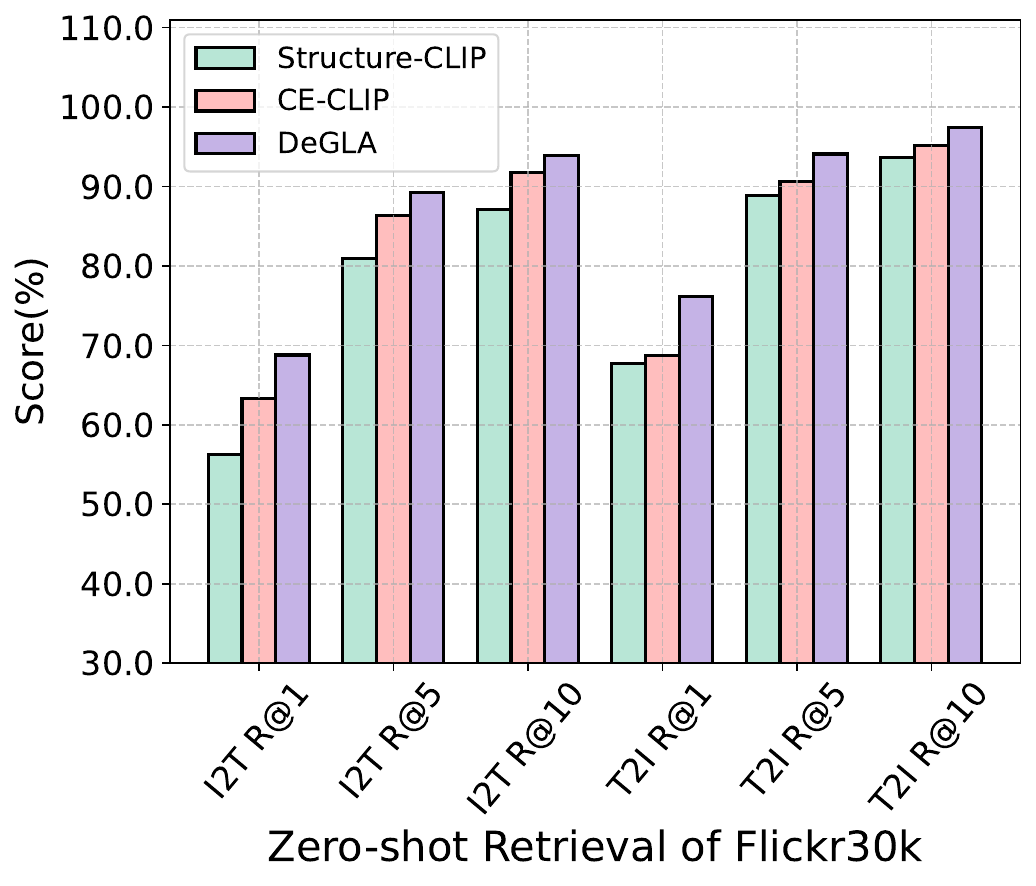} 
    \end{subfigure}
    \vspace{-7mm}
    \caption{Zero-shot image-text retrieval performance comparison on MSCOCO and Flickr30k.}
    \label{fig:retrieval}
    \vspace{-5mm}
\end{figure}

\noindent\textbf{Evaluation Setup.} To comprehensively evaluate DeGLA, we conduct comparative experiments on three compositional reasoning benchmarks: ARO~\cite{yuksekgonul2022and}, VALSE~\cite{parcalabescu2021valse}, and SugarCrepe~\cite{hsieh2024sugarcrepe}. To verify that DeGLA retains general capabilities, we further assess its performance on zero-shot classification, liner probe, and retrieval tasks. For zero-shot classification, we evaluate on eleven datasets: CIFAR-10, CIFAR-100~\cite{zs_cls1}, Food101~\cite{zs_cls2}, Oxford Pets~\cite{zs_cls3}, Flowers102~\cite{zs_cls4}, SUN397~\cite{zs_cls5}, Stanford Cars~\cite{zs_cls6}, DTD~\cite{zs_cls7}, Caltech101~\cite{zs_cls8}, FGVC-Aircraft~\cite{zs_cls9}, and ImageNet~\cite{zs_cls10}.
The datasets used for evaluating the linear probe is the same as that used for zero-shot classification. For zero-shot image-text retrieval, we report results on MSCOCO~\cite{chen2015microsoft} and Flickr30k~\cite{zs_ret_flcker}.\\
We compare DeGLA with three kinds of models: 
(1)Text-only models, including vera~\cite{vera} and Grammar~\cite{Grammar}.
(2) State-of-the-art generative vision-language models, including BLIP~\cite{li2022blip}, BLIP-2~\cite{li2023blip}, MiniGPT-4~\cite{zhu2023minigpt}); (3) High-performance vision-language understanding models, such as BEIT-3~\cite{wang2022image}, XVLM~\cite{zeng2021multi}; (4) Specialized compositional improvement methods NegCLIP~\cite{yuksekgonul2022and}, CyCLIP~\cite{goel2022cyclip}, Structure-CLIP~\cite{huang2024structure}, and CE-CLIP~\cite{zhang2024contrasting}.

\begin{figure}[t!]
    \centering
    \includegraphics[width=0.75\linewidth]{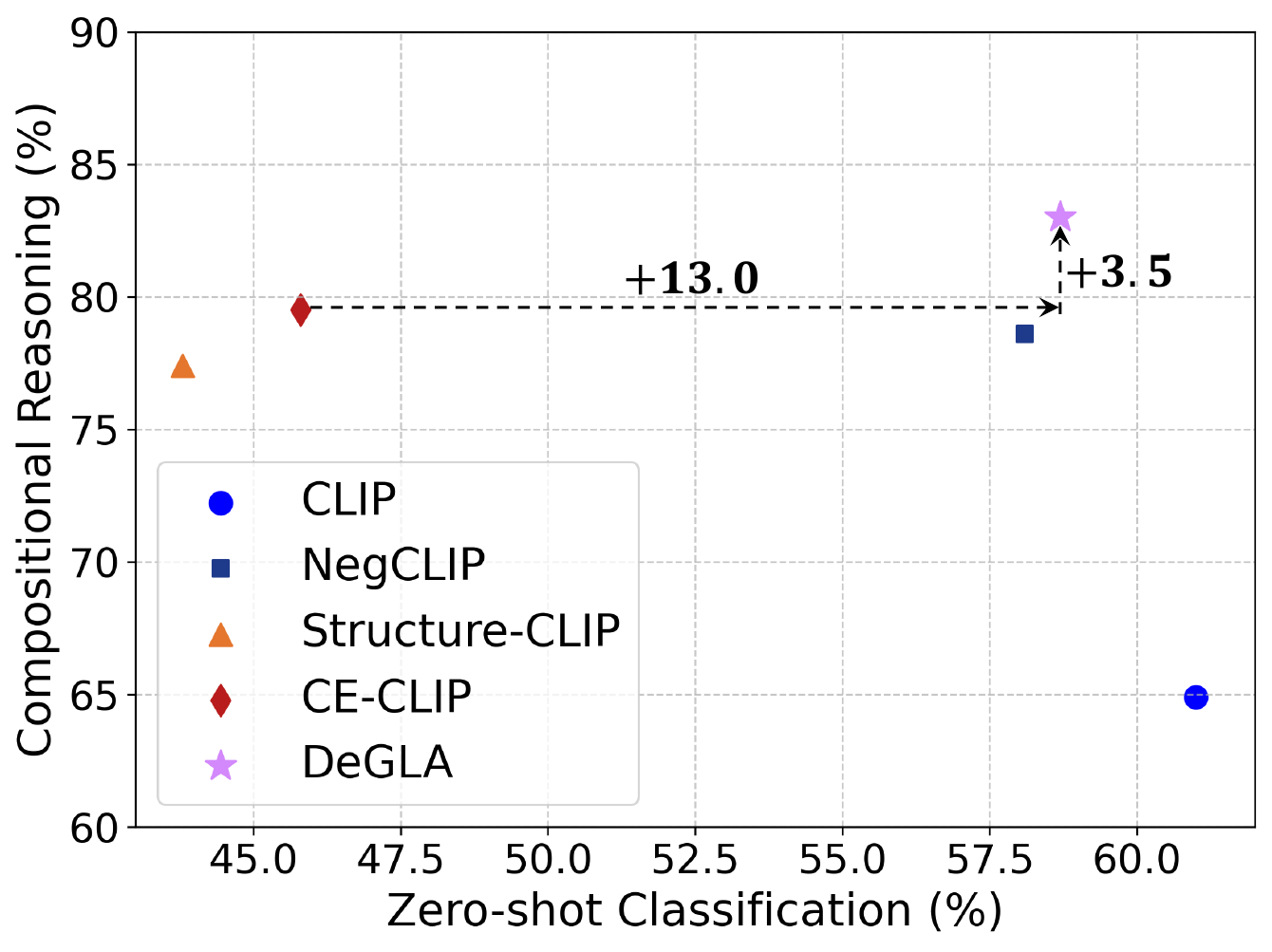}
    \vspace{-3mm}
    \caption{Performance trade off between compositional reasoning~(average performance on VALSE, SugarCrepe, and ARO benchmarks) and zero-shot classification.}
    \label{fig:trade off}
    \vspace{-5mm}
\end{figure}

\begin{figure*}[t!]
\centering

\begin{subfigure}[c]{0.29\textwidth}
    \centering
    \includegraphics[width=\linewidth]{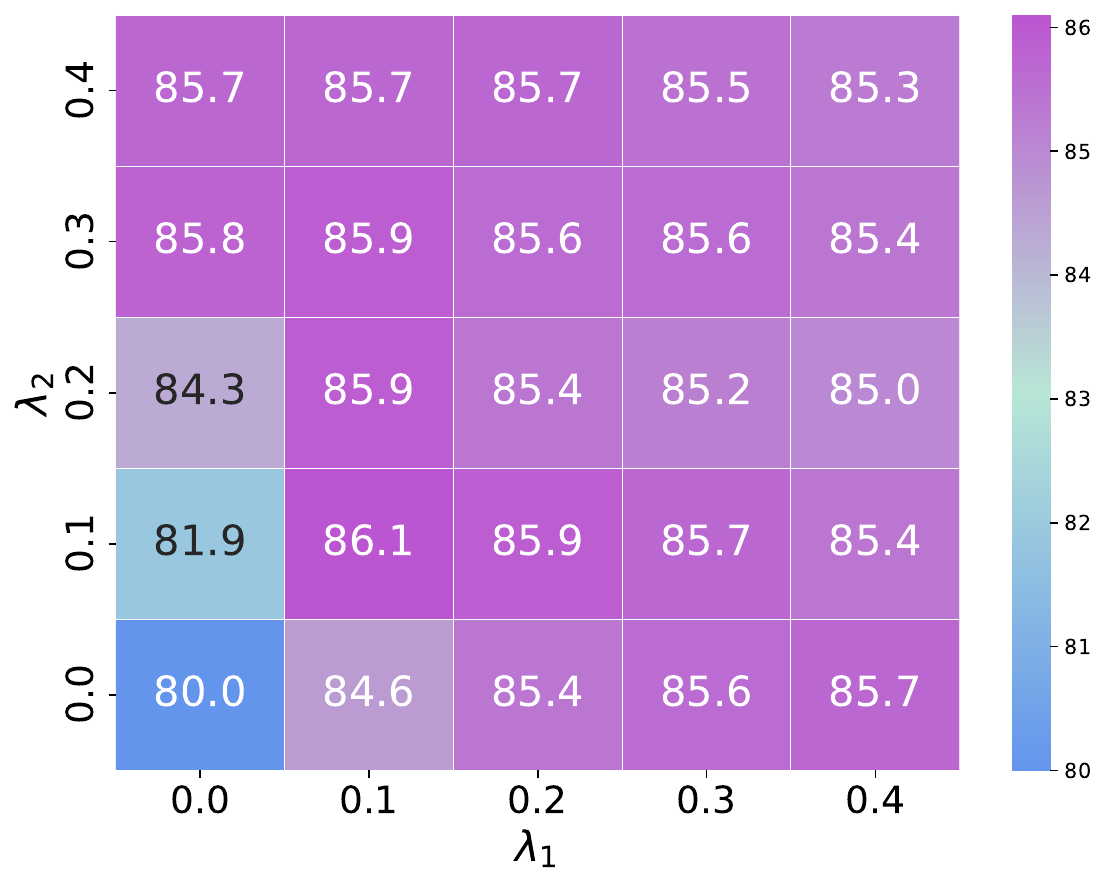}
    \vspace{-5mm}
    \caption{Ablation of $\lambda_1,\lambda_2$}
    \label{fig:ab_l1_l2}
\end{subfigure}
\hfill
\begin{subfigure}[c]{0.31\textwidth}
    \centering
    \includegraphics[width=\linewidth]{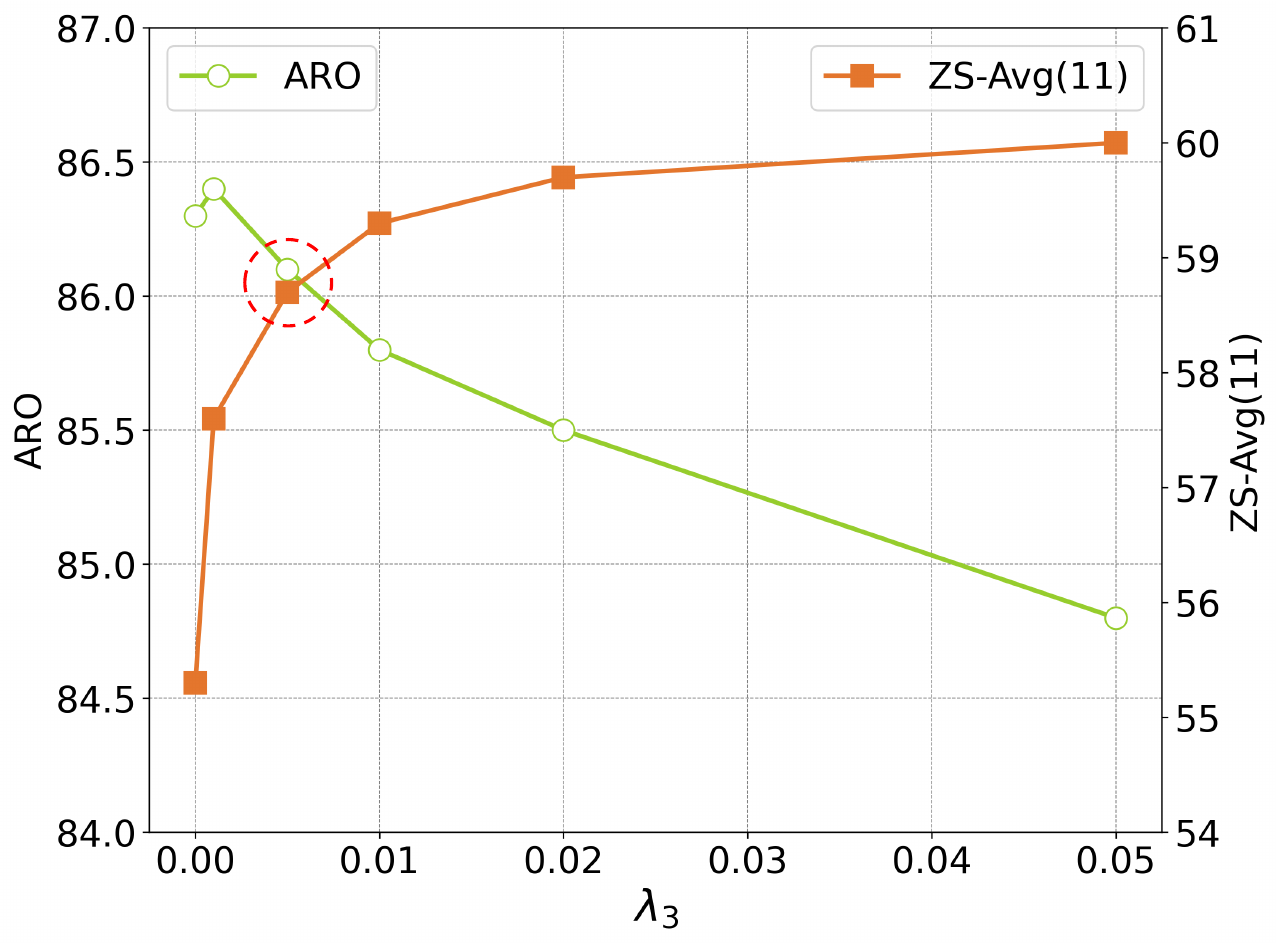}
    \vspace{-5mm}
    \caption{Ablation of $\lambda_3$}
    \label{fig:ab_l3}
\end{subfigure}
\hfill
\begin{subfigure}[c]{0.36\textwidth}
    \centering
    \includegraphics[width=\linewidth]{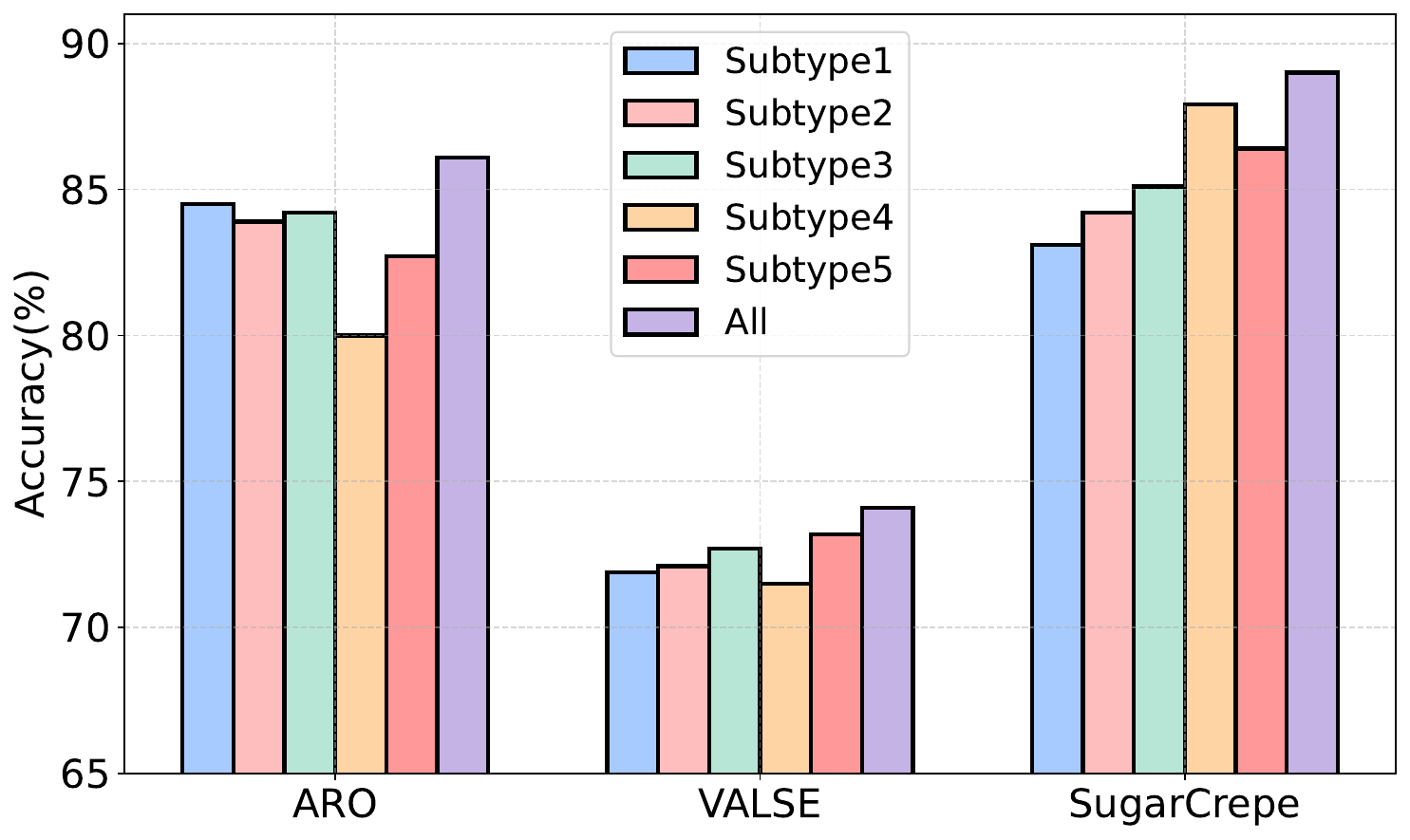}
    \vspace{-5mm}
    \caption{Ablation of hard negative types.}
    \label{fig:negative_types}
\end{subfigure}
\vspace{-3mm}
\caption{Ablation of the different loss weights and hard negative types.}
\label{fig:ablation_all}
\vspace{-3mm}
\end{figure*}

\subsection{Compositional Reasoning}
\label{subsec:main_result}

\noindent\textbf{Performance on VALSE.} We first evaluate DeGLA on VALSE, a benchmark specifically designed to assess pre-trained vision-language models' sensitivity to foiled instances. This benchmark systematically tests fundamental linguistic phenomena across visual and linguistic modalities, including existence, plurality, counting, spatial relations, actions, and entity coreference. As indicated in Table~\ref{tab:valse}, DeGLA outperforms other models in existence, counting, and spatial relations, and ranks second in plurality, actions replacement, coreference standard, and foil-it. Compared to CLIP, DeGLA exhibits an 8.8\% average performance improvement, demonstrating that our approach more effectively leverages hard negatives through local and global alignment strategies. Thanks to the diverse negative samples generated in this study and the employment of both image-grounded and text-grounded contrastive learning, DeGLA significantly surpasses CE-CLIP, achieving an overall performance increase of 1.9\%.

\noindent\textbf{Performance on SugarCrepe.} SugarCrepe is a benchmark designed to reduce language bias in existing benchmarks and provide a more accurate assessment of a model's compositional understanding. As indicated in Table~\ref{tab:sugarcrepe}, DeGLA achieves new state-of-the-art results across all metrics. DeGLA shows significant improvements over the CLIP model, with increases of 10.3\% on Replace, 17.9\% on Swap, and 22.1\% on Add. These results suggest that DeGLA effectively distinguishes between positive and negative samples through local alignment. Compared to CE-CLIP, our method achieves performance enhancements of 3.5\%, 6.9\%, and 1.9\% on Replace, Swap, and Add, respectively. This substantial improvement is attributed to two primary factors: First, the negative sample generation method we introduced leverages Large Language Models (LLMs) to produce high-quality negative text descriptions, reducing the generation of noisy and challenging positive samples. Second, we implement image-grounded and text-grounded contrast mechanisms that enhance the model's discriminative capabilities by drawing positive sample pairs closer and pushing negative sample pairs further apart in the feature space, thus improving the model's compositional understanding.

\noindent\textbf{Performance on ARO.}
We present the performance of our proposed DeGLA on the ARO benchmark in Table~\ref{tab:aro}. DeGLA achieves substantial improvements, registering an average performance increase of 28.7\% over CLIP and 4.9\% over CE-CLIP. It is noteworthy that while DeGLA exhibits significant overall performance gains, it still trails CE-CLIP and Structure-CLIP in the domains of relations and attributes. The primary reason for this discrepancy is that our negative sample generation method prioritizes the production of a diverse array of negative samples, rather than focusing specifically on enhancing the understanding of relations and attributes.

\begin{table}[t!]
\centering 
\caption{Ablation of different components. CN: compositional negatives. SD: self-distillation.}
\label{tab:ablation}
\vspace{-2mm}
\resizebox{\linewidth}{!}{
    \begin{tabular}{c|cccc|cc}
    \toprule
    \textbf{Model} & 
    \textbf{CN} & \textbf{IGC} & \textbf{TGC} & \textbf{SD}  & 
    \textbf{ARO} & 
    \textbf{ZS-Avg.(11)} \\
    \midrule
     \textcolor{gray}{CLIP}~\cite{radford2021learning} & & & & &  \textcolor{gray}{57.4} & \textcolor{gray}{61.0}\\
    \textcolor{gray}{CE-CLIP}~\cite{zhang2024contrasting} & & & & & \textcolor{gray}{81.2} & \textcolor{gray}{45.7}\\
    \midrule
    \multirow{5}{*}{DeGLA(ours)}
     & \ding{51} & & &  & 80.8 & 55.4 \\
     & \ding{51} &\ding{51}& &  & 84.8 & 55.1 \\
     & \ding{51} &&\ding{51} &  & 82.2 & \underline{57.2} \\
     & \ding{51} &\ding{51}& \ding{51}&  &  \textbf{86.2} & 56.9 \\   
     & \ding{51} &\ding{51} & \ding{51}&\ding{51}  &\underline{86.1} & \textbf{58.7} \\
    \bottomrule
\end{tabular}}
\vspace{-3mm}

\end{table}

\subsection{General Understanding}
\noindent\textbf{Zero-shot classification.} 
In Table~\ref{tab:classification}, we present the zero-shot classification performance across 11 datasets. Notably, while Structure-CLIP and CE-CLIP exhibit robust compositional understanding, they significantly reduce the model's original general capabilities, with average accuracies decreasing by 17.2\% and 15.3\%, respectively, compared to CLIP. In contrast, our proposed DeGLA employs a self-distillation constraint module for global alignment, effectively minimizing the loss of general capability. Our method shows a 13.0\% improvement in average accuracy over CE-CLIP, while substantially preserving the model’s inherent general capabilities. As illustrated in Figure~\ref{fig:trade off}, DeGLA achieves a more effective balance between compositional reasoning and general comprehension capabilities compared to other methods.

\noindent\textbf{Linear probe.} 
In Table~\ref{tab:classification}, we detail the linear probe performance across 11 datasets. Consistent with the zero-shot classification results, Structure-CLIP and CE-CLIP significantly reduce the model's original general capabilities, with average accuracies decreasing by 7.8\% and 2.9\%, respectively, compared to CLIP. Conversely, our proposed DeGLA model not only demonstrates superior compositional understanding relative to CE-CLIP but also achieves a 2.3\% average performance improvement in linear probe tasks across these datasets.
\begin{figure*}[t!]
    \centering
    \includegraphics[width=\linewidth]{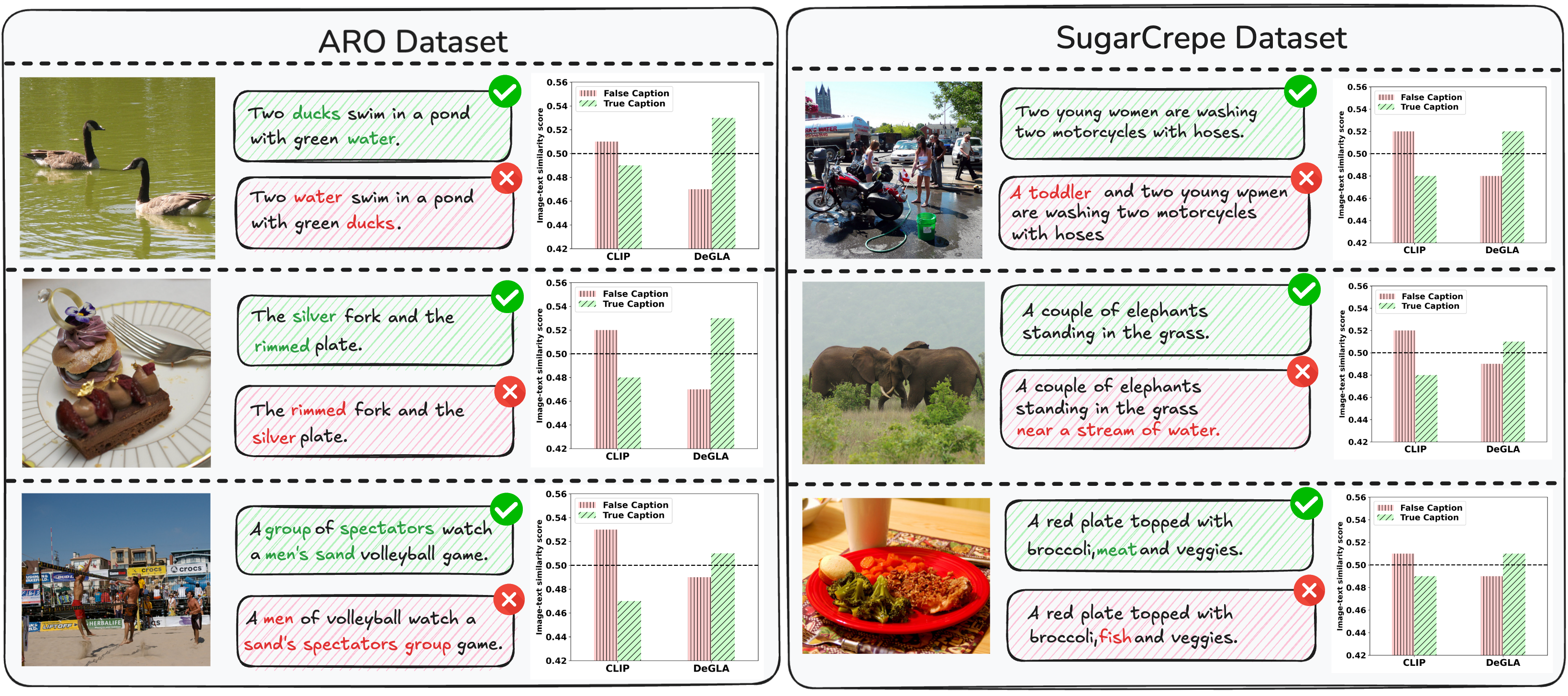}
    \vspace{-7mm}
    \caption{The case study between CLIP and DeGLA. \textcolor{mygreen}{Green \ding{52}} indicates true caption, while \textcolor{red}{red \ding{55}} indicates false caption. The bar chart represents the model's prediction results.}
    \vspace{-3mm}
    \label{fig:case_vis}
\end{figure*}

\noindent\textbf{Zero-shot image text retrieval.}
We compare the zero-shot retrieval performance of DeGLA, Structure CLIP, and CE-CLIP on the MSCOCO and Flickr30k datasets.  As illustrated in Figure~\ref{fig:retrieval}, DeGLA surpasses previous models on both datasets, demonstrating superior performance in standard retrieval tasks. This finding is consistent with our earlier experimental results, further substantiating DeGLA's excellence in both general capabilities and compositional reasoning.

\subsection{Analysis}
\label{subsec:ablation}
\noindent\textbf{Trade-off Analysis.}
As illustrated in Figure~\ref{fig:trade off}, we analyze the trade-off between general capabilities and compositional reasoning. For the evaluation of compositional reasoning, we utilize the average scores from three benchmarks. Typically, enhancements in compositional reasoning are accompanied by declines in general capabilities, as demonstrated by CE-CLIP, which, despite significant gains in compositional reasoning compared to CLIP, suffers a substantial reduction in general capabilities. In contrast, through its meticulously designed data generation pipeline and training framework, DeGLA achieves an optimal balance. Compared to the baseline CE-CLIP, DeGLA records a 13\% improvement in zero-shot classification and an average enhancement of 3.5\% in compositional reasoning tasks, further underscoring its superiority.

\noindent\textbf{Ablation of different components.}
To validate the efficacy of the proposed method, we perform comprehensive ablation studies on key components using the ARO benchmark in Table~\ref{tab:ablation}. By combining all modules, DeGLA achieves the most balanced performance, attaining an 86.1\% average score on ARO with 4.9\% higher than the previous SOTA (CE-CLIP). For zero-shot classification, it outperforms CE-CLIP by 13.0\%, demonstrating superior generalization. Analysis of component contributions reveals that integrating high-quality negative samples increases ARO performance by 23.4\% but decreases zero-shot classification performance by 5.6\%. The local alignment losses, IGC and TGC, improve ARO performance by 4.0\% and 1.4\%, respectively. After Combining further boosting ARO performance by 5.4\%, while zero-shot classification still decreases by 4.1\%. Notably, TGC mitigates declines in general capability by using frozen text embeddings from the EMA text encoder, ensuring minimal deviation from the pretrained CLIP model during training. The introduction of a self-distillation mechanism significantly enhances general capability, improving zero-shot classification by 1.8\% with a negligible 0.1\% loss in ARO performance.


\noindent\textbf{Ablation of compositional negative texts.}
To verify the effectiveness of our proposed instruction-based negative generation pipeline, we conduct an ablation analysis on five subtypes of negative texts, as depicted in Figure~\ref{fig:negative_types}. The results show that combining these subtypes achieves optimal performance, thereby validating the pipeline's effectiveness. This success is attributed to the representativeness of our samples, which effectively enhance the compositional understanding of CLIP.

\subsection{Case Study}

Figure~\ref{fig:case_vis} compares CLIP and DeGLA on ARO and SugarCrepe. Consistent with NegCLIP~\cite{yuksekgonul2022and}, CLIP shows a “bag-of-words” issue, struggling with structural text changes like word order swaps or substitutions—for example, failing to distinguish between “Two ducks swim in a pond with green water” and “Two swans swim in a pond with green ducks.” In contrast, DeGLA correctly differentiates these based on the image. After training on our compositionality-focused hard negatives (Section~\ref{sec:negative data}), DeGLA demonstrates improved compositional reasoning over pretrained CLIP, validating the effectiveness of our data generation pipeline.
\section{Conclusion}
In this paper, we observe that while previous methods enhance CLIP’s compositional understanding, they often compromise its general capabilities. To overcome this limitation, we introduce a \textbf{De}coupled \textbf{G}lobal-\textbf{L}ocal \textbf{A}lignment~(\textbf{DeGLA}) framework, which not only improves compositional understanding but also significantly reduces losses in general capabilities. To optimize the retention of the model’s inherent capabilities, we incorporate a self-distillation mechanism within the global alignment process, aligning the learnable image-text encoder with a frozen teacher model derived from an exponential moving average. For enhancing compositional understanding, we leverage the in-context learning capabilities of Large Language Models (LLMs) to generate approximately 2M high-quality negative captions across five types. We further propose the Image-Grounded Contrast (IGC) loss and Text-Grounded Contrast (TGC) loss to improve vision-language compositionality.
\begin{acks}
This work is supported by National Natural Science Foundation of China (Grant No.
62301046)
\end{acks}

\bibliographystyle{ACM-Reference-Format}
\balance
\bibliography{main}


\begin{thebibliography}{70}


\ifx \showCODEN    \undefined \def \showCODEN     #1{\unskip}     \fi
\ifx \showISBNx    \undefined \def \showISBNx     #1{\unskip}     \fi
\ifx \showISBNxiii \undefined \def \showISBNxiii  #1{\unskip}     \fi
\ifx \showISSN     \undefined \def \showISSN      #1{\unskip}     \fi
\ifx \showLCCN     \undefined \def \showLCCN      #1{\unskip}     \fi
\ifx \shownote     \undefined \def \shownote      #1{#1}          \fi
\ifx \showarticletitle \undefined \def \showarticletitle #1{#1}   \fi
\ifx \showURL      \undefined \def \showURL       {\relax}        \fi
\providecommand\bibfield[2]{#2}
\providecommand\bibinfo[2]{#2}
\providecommand\natexlab[1]{#1}
\providecommand\showeprint[2][]{arXiv:#2}

\bibitem[Alayrac et~al\mbox{.}(2022)]%
        {alayrac2022flamingo}
\bibfield{author}{\bibinfo{person}{Jean-Baptiste Alayrac}, \bibinfo{person}{Jeff Donahue}, \bibinfo{person}{Pauline Luc}, \bibinfo{person}{Antoine Miech}, \bibinfo{person}{Iain Barr}, \bibinfo{person}{Yana Hasson}, \bibinfo{person}{Karel Lenc}, \bibinfo{person}{Arthur Mensch}, \bibinfo{person}{Katherine Millican}, \bibinfo{person}{Malcolm Reynolds}, {et~al\mbox{.}}} \bibinfo{year}{2022}\natexlab{}.
\newblock \showarticletitle{Flamingo: a visual language model for few-shot learning}.
\newblock \bibinfo{journal}{\emph{NIPS}}  \bibinfo{volume}{35} (\bibinfo{year}{2022}), \bibinfo{pages}{23716--23736}.
\newblock


\bibitem[Bossard et~al\mbox{.}(2014)]%
        {zs_cls2}
\bibfield{author}{\bibinfo{person}{Lukas Bossard}, \bibinfo{person}{Matthieu Guillaumin}, {and} \bibinfo{person}{Luc Van~Gool}.} \bibinfo{year}{2014}\natexlab{}.
\newblock \showarticletitle{Food-101--mining discriminative components with random forests}. In \bibinfo{booktitle}{\emph{Computer vision--ECCV 2014: 13th European conference, zurich, Switzerland, September 6-12, 2014, proceedings, part VI 13}}. Springer, \bibinfo{pages}{446--461}.
\newblock


\bibitem[Chen et~al\mbox{.}(2015)]%
        {chen2015microsoft}
\bibfield{author}{\bibinfo{person}{Xinlei Chen}, \bibinfo{person}{Hao Fang}, \bibinfo{person}{Tsung-Yi Lin}, \bibinfo{person}{Ramakrishna Vedantam}, \bibinfo{person}{Saurabh Gupta}, \bibinfo{person}{Piotr Doll{\'a}r}, {and} \bibinfo{person}{C~Lawrence Zitnick}.} \bibinfo{year}{2015}\natexlab{}.
\newblock \showarticletitle{Microsoft coco captions: Data collection and evaluation server}.
\newblock \bibinfo{journal}{\emph{arXiv preprint arXiv:1504.00325}} (\bibinfo{year}{2015}).
\newblock


\bibitem[Cho et~al\mbox{.}(2024)]%
        {cho2024cat}
\bibfield{author}{\bibinfo{person}{Seokju Cho}, \bibinfo{person}{Heeseong Shin}, \bibinfo{person}{Sunghwan Hong}, \bibinfo{person}{Anurag Arnab}, \bibinfo{person}{Paul~Hongsuck Seo}, {and} \bibinfo{person}{Seungryong Kim}.} \bibinfo{year}{2024}\natexlab{}.
\newblock \showarticletitle{Cat-seg: Cost aggregation for open-vocabulary semantic segmentation}. In \bibinfo{booktitle}{\emph{CVPR}}. \bibinfo{pages}{4113--4123}.
\newblock


\bibitem[Cimpoi et~al\mbox{.}(2014)]%
        {zs_cls7}
\bibfield{author}{\bibinfo{person}{Mircea Cimpoi}, \bibinfo{person}{Subhransu Maji}, \bibinfo{person}{Iasonas Kokkinos}, \bibinfo{person}{Sammy Mohamed}, {and} \bibinfo{person}{Andrea Vedaldi}.} \bibinfo{year}{2014}\natexlab{}.
\newblock \showarticletitle{Describing textures in the wild}. In \bibinfo{booktitle}{\emph{CVPR}}. \bibinfo{pages}{3606--3613}.
\newblock


\bibitem[Deng et~al\mbox{.}(2009)]%
        {zs_cls10}
\bibfield{author}{\bibinfo{person}{Jia Deng}, \bibinfo{person}{Wei Dong}, \bibinfo{person}{Richard Socher}, \bibinfo{person}{Li-Jia Li}, \bibinfo{person}{Kai Li}, {and} \bibinfo{person}{Li Fei-Fei}.} \bibinfo{year}{2009}\natexlab{}.
\newblock \showarticletitle{Imagenet: A large-scale hierarchical image database}. In \bibinfo{booktitle}{\emph{CVPR}}. Ieee, \bibinfo{pages}{248--255}.
\newblock


\bibitem[Doveh et~al\mbox{.}(2023a)]%
        {doveh2023dense}
\bibfield{author}{\bibinfo{person}{Sivan Doveh}, \bibinfo{person}{Assaf Arbelle}, \bibinfo{person}{Sivan Harary}, \bibinfo{person}{Roei Herzig}, \bibinfo{person}{Donghyun Kim}, \bibinfo{person}{Paola Cascante-Bonilla}, \bibinfo{person}{Amit Alfassy}, \bibinfo{person}{Rameswar Panda}, \bibinfo{person}{Raja Giryes}, \bibinfo{person}{Rogerio Feris}, {et~al\mbox{.}}} \bibinfo{year}{2023}\natexlab{a}.
\newblock \showarticletitle{Dense and aligned captions (dac) promote compositional reasoning in vl models}.
\newblock \bibinfo{journal}{\emph{NIPS}}  \bibinfo{volume}{36} (\bibinfo{year}{2023}), \bibinfo{pages}{76137--76150}.
\newblock


\bibitem[Doveh et~al\mbox{.}(2023b)]%
        {doveh2023teaching}
\bibfield{author}{\bibinfo{person}{Sivan Doveh}, \bibinfo{person}{Assaf Arbelle}, \bibinfo{person}{Sivan Harary}, \bibinfo{person}{Eli Schwartz}, \bibinfo{person}{Roei Herzig}, \bibinfo{person}{Raja Giryes}, \bibinfo{person}{Rogerio Feris}, \bibinfo{person}{Rameswar Panda}, \bibinfo{person}{Shimon Ullman}, {and} \bibinfo{person}{Leonid Karlinsky}.} \bibinfo{year}{2023}\natexlab{b}.
\newblock \showarticletitle{Teaching structured vision \& language concepts to vision \& language models}. In \bibinfo{booktitle}{\emph{CVPR}}. \bibinfo{pages}{2657--2668}.
\newblock


\bibitem[Fan et~al\mbox{.}(2023)]%
        {fan2023improving}
\bibfield{author}{\bibinfo{person}{Lijie Fan}, \bibinfo{person}{Dilip Krishnan}, \bibinfo{person}{Phillip Isola}, \bibinfo{person}{Dina Katabi}, {and} \bibinfo{person}{Yonglong Tian}.} \bibinfo{year}{2023}\natexlab{}.
\newblock \showarticletitle{Improving clip training with language rewrites}.
\newblock \bibinfo{journal}{\emph{NIPS}}  \bibinfo{volume}{36} (\bibinfo{year}{2023}), \bibinfo{pages}{35544--35575}.
\newblock


\bibitem[Fei-Fei et~al\mbox{.}(2004)]%
        {zs_cls8}
\bibfield{author}{\bibinfo{person}{Li Fei-Fei}, \bibinfo{person}{Rob Fergus}, {and} \bibinfo{person}{Pietro Perona}.} \bibinfo{year}{2004}\natexlab{}.
\newblock \showarticletitle{Learning generative visual models from few training examples: An incremental bayesian approach tested on 101 object categories}. In \bibinfo{booktitle}{\emph{CVPR}}. IEEE, \bibinfo{pages}{178--178}.
\newblock


\bibitem[Geng et~al\mbox{.}(2023)]%
        {geng2023hiclip}
\bibfield{author}{\bibinfo{person}{Shijie Geng}, \bibinfo{person}{Jianbo Yuan}, \bibinfo{person}{Yu Tian}, \bibinfo{person}{Yuxiao Chen}, {and} \bibinfo{person}{Yongfeng Zhang}.} \bibinfo{year}{2023}\natexlab{}.
\newblock \showarticletitle{HiCLIP: Contrastive language-image pretraining with hierarchy-aware attention}.
\newblock \bibinfo{journal}{\emph{ICLR}} (\bibinfo{year}{2023}).
\newblock


\bibitem[Goel et~al\mbox{.}(2022)]%
        {goel2022cyclip}
\bibfield{author}{\bibinfo{person}{Shashank Goel}, \bibinfo{person}{Hritik Bansal}, \bibinfo{person}{Sumit Bhatia}, \bibinfo{person}{Ryan Rossi}, \bibinfo{person}{Vishwa Vinay}, {and} \bibinfo{person}{Aditya Grover}.} \bibinfo{year}{2022}\natexlab{}.
\newblock \showarticletitle{Cyclip: Cyclic contrastive language-image pretraining}.
\newblock \bibinfo{journal}{\emph{NIPS}}  \bibinfo{volume}{35} (\bibinfo{year}{2022}), \bibinfo{pages}{6704--6719}.
\newblock


\bibitem[Grattafiori et~al\mbox{.}(2024)]%
        {llama3.1}
\bibfield{author}{\bibinfo{person}{Aaron Grattafiori}, \bibinfo{person}{Abhimanyu Dubey}, \bibinfo{person}{Abhinav Jauhri}, \bibinfo{person}{Abhinav Pandey}, \bibinfo{person}{Abhishek Kadian}, \bibinfo{person}{Ahmad Al-Dahle}, \bibinfo{person}{Aiesha Letman}, \bibinfo{person}{Akhil Mathur}, \bibinfo{person}{Alan Schelten}, \bibinfo{person}{Alex Vaughan}, {et~al\mbox{.}}} \bibinfo{year}{2024}\natexlab{}.
\newblock \showarticletitle{The llama 3 herd of models}.
\newblock \bibinfo{journal}{\emph{arXiv preprint arXiv:2407.21783}} (\bibinfo{year}{2024}).
\newblock


\bibitem[Gu et~al\mbox{.}(2024)]%
        {rwkvclip}
\bibfield{author}{\bibinfo{person}{Tiancheng Gu}, \bibinfo{person}{Kaicheng Yang}, \bibinfo{person}{Xiang An}, \bibinfo{person}{Ziyong Feng}, \bibinfo{person}{Dongnan Liu}, \bibinfo{person}{Weidong Cai}, {and} \bibinfo{person}{Jiankang Deng}.} \bibinfo{year}{2024}\natexlab{}.
\newblock \showarticletitle{RWKV-CLIP: a robust vision-language representation learner}.
\newblock \bibinfo{journal}{\emph{arXiv preprint arXiv:2406.06973}} (\bibinfo{year}{2024}).
\newblock


\bibitem[Gu et~al\mbox{.}(2025a)]%
        {gu2025breaking}
\bibfield{author}{\bibinfo{person}{Tiancheng Gu}, \bibinfo{person}{Kaicheng Yang}, \bibinfo{person}{Ziyong Feng}, \bibinfo{person}{Xingjun Wang}, \bibinfo{person}{Yanzhao Zhang}, \bibinfo{person}{Dingkun Long}, \bibinfo{person}{Yingda Chen}, \bibinfo{person}{Weidong Cai}, {and} \bibinfo{person}{Jiankang Deng}.} \bibinfo{year}{2025}\natexlab{a}.
\newblock \showarticletitle{Breaking the Modality Barrier: Universal Embedding Learning with Multimodal LLMs}.
\newblock \bibinfo{journal}{\emph{arXiv preprint arXiv:2504.17432}} (\bibinfo{year}{2025}).
\newblock


\bibitem[Gu et~al\mbox{.}(2025b)]%
        {gu2025realsyn}
\bibfield{author}{\bibinfo{person}{Tiancheng Gu}, \bibinfo{person}{Kaicheng Yang}, \bibinfo{person}{Chaoyi Zhang}, \bibinfo{person}{Yin Xie}, \bibinfo{person}{Xiang An}, \bibinfo{person}{Ziyong Feng}, \bibinfo{person}{Dongnan Liu}, \bibinfo{person}{Weidong Cai}, {and} \bibinfo{person}{Jiankang Deng}.} \bibinfo{year}{2025}\natexlab{b}.
\newblock \showarticletitle{RealSyn: An Effective and Scalable Multimodal Interleaved Document Transformation Paradigm}.
\newblock \bibinfo{journal}{\emph{arXiv preprint arXiv:2502.12513}} (\bibinfo{year}{2025}).
\newblock


\bibitem[Gu et~al\mbox{.}(2021)]%
        {gu2021open}
\bibfield{author}{\bibinfo{person}{Xiuye Gu}, \bibinfo{person}{Tsung-Yi Lin}, \bibinfo{person}{Weicheng Kuo}, {and} \bibinfo{person}{Yin Cui}.} \bibinfo{year}{2021}\natexlab{}.
\newblock \showarticletitle{Open-vocabulary object detection via vision and language knowledge distillation}.
\newblock \bibinfo{journal}{\emph{arXiv preprint arXiv:2104.13921}} (\bibinfo{year}{2021}).
\newblock


\bibitem[Hinton(2015)]%
        {hinton2015distilling}
\bibfield{author}{\bibinfo{person}{Geoffrey Hinton}.} \bibinfo{year}{2015}\natexlab{}.
\newblock \showarticletitle{Distilling the Knowledge in a Neural Network}.
\newblock \bibinfo{journal}{\emph{arXiv preprint arXiv:1503.02531}} (\bibinfo{year}{2015}).
\newblock


\bibitem[Hoyer et~al\mbox{.}(2025)]%
        {hoyer2025semivl}
\bibfield{author}{\bibinfo{person}{Lukas Hoyer}, \bibinfo{person}{David~Joseph Tan}, \bibinfo{person}{Muhammad~Ferjad Naeem}, \bibinfo{person}{Luc Van~Gool}, {and} \bibinfo{person}{Federico Tombari}.} \bibinfo{year}{2025}\natexlab{}.
\newblock \showarticletitle{Semivl: Semi-supervised semantic segmentation with vision-language guidance}. In \bibinfo{booktitle}{\emph{ECCV}}. Springer, \bibinfo{pages}{257--275}.
\newblock


\bibitem[Hsieh et~al\mbox{.}(2024)]%
        {hsieh2024sugarcrepe}
\bibfield{author}{\bibinfo{person}{Cheng-Yu Hsieh}, \bibinfo{person}{Jieyu Zhang}, \bibinfo{person}{Zixian Ma}, \bibinfo{person}{Aniruddha Kembhavi}, {and} \bibinfo{person}{Ranjay Krishna}.} \bibinfo{year}{2024}\natexlab{}.
\newblock \showarticletitle{Sugarcrepe: Fixing hackable benchmarks for vision-language compositionality}.
\newblock \bibinfo{journal}{\emph{NIPS}}  \bibinfo{volume}{36} (\bibinfo{year}{2024}).
\newblock


\bibitem[Huang et~al\mbox{.}(2024b)]%
        {huang2024etag}
\bibfield{author}{\bibinfo{person}{Libo Huang}, \bibinfo{person}{Yan Zeng}, \bibinfo{person}{Chuanguang Yang}, \bibinfo{person}{Zhulin An}, \bibinfo{person}{Boyu Diao}, {and} \bibinfo{person}{Yongjun Xu}.} \bibinfo{year}{2024}\natexlab{b}.
\newblock \showarticletitle{eTag: Class-Incremental Learning via Embedding Distillation and Task-Oriented Generation}. In \bibinfo{booktitle}{\emph{AAAI}}, Vol.~\bibinfo{volume}{38}. \bibinfo{pages}{12591--12599}.
\newblock


\bibitem[Huang et~al\mbox{.}(2024a)]%
        {huang2024structure}
\bibfield{author}{\bibinfo{person}{Yufeng Huang}, \bibinfo{person}{Jiji Tang}, \bibinfo{person}{Zhuo Chen}, \bibinfo{person}{Rongsheng Zhang}, \bibinfo{person}{Xinfeng Zhang}, \bibinfo{person}{Weijie Chen}, \bibinfo{person}{Zeng Zhao}, \bibinfo{person}{Zhou Zhao}, \bibinfo{person}{Tangjie Lv}, \bibinfo{person}{Zhipeng Hu}, {et~al\mbox{.}}} \bibinfo{year}{2024}\natexlab{a}.
\newblock \showarticletitle{Structure-CLIP: Towards Scene Graph Knowledge to Enhance Multi-Modal Structured Representations}. In \bibinfo{booktitle}{\emph{AAAI}}, Vol.~\bibinfo{volume}{38}. \bibinfo{pages}{2417--2425}.
\newblock


\bibitem[Jiao et~al\mbox{.}(2019)]%
        {jiao2019tinybert}
\bibfield{author}{\bibinfo{person}{Xiaoqi Jiao}, \bibinfo{person}{Yichun Yin}, \bibinfo{person}{Lifeng Shang}, \bibinfo{person}{Xin Jiang}, \bibinfo{person}{Xiao Chen}, \bibinfo{person}{Linlin Li}, \bibinfo{person}{Fang Wang}, {and} \bibinfo{person}{Qun Liu}.} \bibinfo{year}{2019}\natexlab{}.
\newblock \showarticletitle{Tinybert: Distilling bert for natural language understanding}.
\newblock \bibinfo{journal}{\emph{arXiv preprint arXiv:1909.10351}} (\bibinfo{year}{2019}).
\newblock


\bibitem[Kamath et~al\mbox{.}(2023)]%
        {kamath2023s}
\bibfield{author}{\bibinfo{person}{Amita Kamath}, \bibinfo{person}{Jack Hessel}, {and} \bibinfo{person}{Kai-Wei Chang}.} \bibinfo{year}{2023}\natexlab{}.
\newblock \showarticletitle{What's" up" with vision-language models? Investigating their struggle with spatial reasoning}.
\newblock \bibinfo{journal}{\emph{arXiv preprint arXiv:2310.19785}} (\bibinfo{year}{2023}).
\newblock


\bibitem[Kamath et~al\mbox{.}(2024)]%
        {hard_positive}
\bibfield{author}{\bibinfo{person}{Amita Kamath}, \bibinfo{person}{Cheng-Yu Hsieh}, \bibinfo{person}{Kai-Wei Chang}, {and} \bibinfo{person}{Ranjay Krishna}.} \bibinfo{year}{2024}\natexlab{}.
\newblock \showarticletitle{The hard positive truth about vision-language compositionality}. In \bibinfo{booktitle}{\emph{ECCV}}. Springer, \bibinfo{pages}{37--54}.
\newblock


\bibitem[Karpathy and Fei-Fei(2015)]%
        {karpathy2015deep}
\bibfield{author}{\bibinfo{person}{Andrej Karpathy} {and} \bibinfo{person}{Li Fei-Fei}.} \bibinfo{year}{2015}\natexlab{}.
\newblock \showarticletitle{Deep visual-semantic alignments for generating image descriptions}. In \bibinfo{booktitle}{\emph{CVPR}}. \bibinfo{pages}{3128--3137}.
\newblock


\bibitem[Kaul et~al\mbox{.}(2023)]%
        {object_1}
\bibfield{author}{\bibinfo{person}{Prannay Kaul}, \bibinfo{person}{Weidi Xie}, {and} \bibinfo{person}{Andrew Zisserman}.} \bibinfo{year}{2023}\natexlab{}.
\newblock \showarticletitle{Multi-modal classifiers for open-vocabulary object detection}. In \bibinfo{booktitle}{\emph{ICML}}. PMLR, \bibinfo{pages}{15946--15969}.
\newblock


\bibitem[Krause et~al\mbox{.}(2013)]%
        {zs_cls6}
\bibfield{author}{\bibinfo{person}{Jonathan Krause}, \bibinfo{person}{Michael Stark}, \bibinfo{person}{Jia Deng}, {and} \bibinfo{person}{Li Fei-Fei}.} \bibinfo{year}{2013}\natexlab{}.
\newblock \showarticletitle{3d object representations for fine-grained categorization}. In \bibinfo{booktitle}{\emph{ICCV}}. \bibinfo{pages}{554--561}.
\newblock


\bibitem[Krizhevsky et~al\mbox{.}(2009)]%
        {zs_cls1}
\bibfield{author}{\bibinfo{person}{Alex Krizhevsky}, \bibinfo{person}{Geoffrey Hinton}, {et~al\mbox{.}}} \bibinfo{year}{2009}\natexlab{}.
\newblock \showarticletitle{Learning multiple layers of features from tiny images}.
\newblock  (\bibinfo{year}{2009}).
\newblock


\bibitem[Lee et~al\mbox{.}(2022)]%
        {lee2022uniclip}
\bibfield{author}{\bibinfo{person}{Janghyeon Lee}, \bibinfo{person}{Jongsuk Kim}, \bibinfo{person}{Hyounguk Shon}, \bibinfo{person}{Bumsoo Kim}, \bibinfo{person}{Seung~Hwan Kim}, \bibinfo{person}{Honglak Lee}, {and} \bibinfo{person}{Junmo Kim}.} \bibinfo{year}{2022}\natexlab{}.
\newblock \showarticletitle{Uniclip: Unified framework for contrastive language-image pre-training}.
\newblock \bibinfo{journal}{\emph{NIPS}}  \bibinfo{volume}{35} (\bibinfo{year}{2022}), \bibinfo{pages}{1008--1019}.
\newblock


\bibitem[Li et~al\mbox{.}(2022b)]%
        {li2022language}
\bibfield{author}{\bibinfo{person}{Boyi Li}, \bibinfo{person}{Kilian~Q Weinberger}, \bibinfo{person}{Serge Belongie}, \bibinfo{person}{Vladlen Koltun}, {and} \bibinfo{person}{Ren{\'e} Ranftl}.} \bibinfo{year}{2022}\natexlab{b}.
\newblock \showarticletitle{Language-driven semantic segmentation}.
\newblock \bibinfo{journal}{\emph{arXiv preprint arXiv:2201.03546}} (\bibinfo{year}{2022}).
\newblock


\bibitem[Li et~al\mbox{.}(2023b)]%
        {li2023blip}
\bibfield{author}{\bibinfo{person}{Junnan Li}, \bibinfo{person}{Dongxu Li}, \bibinfo{person}{Silvio Savarese}, {and} \bibinfo{person}{Steven Hoi}.} \bibinfo{year}{2023}\natexlab{b}.
\newblock \showarticletitle{Blip-2: Bootstrapping language-image pre-training with frozen image encoders and large language models}. In \bibinfo{booktitle}{\emph{ICML}}. PMLR, \bibinfo{pages}{19730--19742}.
\newblock


\bibitem[Li et~al\mbox{.}(2022a)]%
        {li2022blip}
\bibfield{author}{\bibinfo{person}{Junnan Li}, \bibinfo{person}{Dongxu Li}, \bibinfo{person}{Caiming Xiong}, {and} \bibinfo{person}{Steven Hoi}.} \bibinfo{year}{2022}\natexlab{a}.
\newblock \showarticletitle{Blip: Bootstrapping language-image pre-training for unified vision-language understanding and generation}. In \bibinfo{booktitle}{\emph{ICML}}. PMLR, \bibinfo{pages}{12888--12900}.
\newblock


\bibitem[Li et~al\mbox{.}(2023a)]%
        {flip}
\bibfield{author}{\bibinfo{person}{Yanghao Li}, \bibinfo{person}{Haoqi Fan}, \bibinfo{person}{Ronghang Hu}, \bibinfo{person}{Christoph Feichtenhofer}, {and} \bibinfo{person}{Kaiming He}.} \bibinfo{year}{2023}\natexlab{a}.
\newblock \showarticletitle{Scaling language-image pre-training via masking}. In \bibinfo{booktitle}{\emph{CVPR}}. \bibinfo{pages}{23390--23400}.
\newblock


\bibitem[Li et~al\mbox{.}(2023c)]%
        {li2023curriculum}
\bibfield{author}{\bibinfo{person}{Zheng Li}, \bibinfo{person}{Xiang Li}, \bibinfo{person}{Lingfeng Yang}, \bibinfo{person}{Borui Zhao}, \bibinfo{person}{Renjie Song}, \bibinfo{person}{Lei Luo}, \bibinfo{person}{Jun Li}, {and} \bibinfo{person}{Jian Yang}.} \bibinfo{year}{2023}\natexlab{c}.
\newblock \showarticletitle{Curriculum temperature for knowledge distillation}. In \bibinfo{booktitle}{\emph{AAAI}}, Vol.~\bibinfo{volume}{37}. \bibinfo{pages}{1504--1512}.
\newblock


\bibitem[Li et~al\mbox{.}(2021)]%
        {li2021online}
\bibfield{author}{\bibinfo{person}{Zheng Li}, \bibinfo{person}{Jingwen Ye}, \bibinfo{person}{Mingli Song}, \bibinfo{person}{Ying Huang}, {and} \bibinfo{person}{Zhigeng Pan}.} \bibinfo{year}{2021}\natexlab{}.
\newblock \showarticletitle{Online knowledge distillation for efficient pose estimation}. In \bibinfo{booktitle}{\emph{ICCV}}. \bibinfo{pages}{11740--11750}.
\newblock


\bibitem[Liu et~al\mbox{.}(2023)]%
        {vera}
\bibfield{author}{\bibinfo{person}{Jiacheng Liu}, \bibinfo{person}{Wenya Wang}, \bibinfo{person}{Dianzhuo Wang}, \bibinfo{person}{Noah~A Smith}, \bibinfo{person}{Yejin Choi}, {and} \bibinfo{person}{Hannaneh Hajishirzi}.} \bibinfo{year}{2023}\natexlab{}.
\newblock \showarticletitle{Vera: A general-purpose plausibility estimation model for commonsense statements}.
\newblock \bibinfo{journal}{\emph{arXiv preprint arXiv:2305.03695}} (\bibinfo{year}{2023}).
\newblock


\bibitem[Ma et~al\mbox{.}(2023)]%
        {object_2}
\bibfield{author}{\bibinfo{person}{Chuofan Ma}, \bibinfo{person}{Yi Jiang}, \bibinfo{person}{Xin Wen}, \bibinfo{person}{Zehuan Yuan}, {and} \bibinfo{person}{Xiaojuan Qi}.} \bibinfo{year}{2023}\natexlab{}.
\newblock \showarticletitle{Codet: Co-occurrence guided region-word alignment for open-vocabulary object detection}.
\newblock \bibinfo{journal}{\emph{NIPS}}  \bibinfo{volume}{36} (\bibinfo{year}{2023}), \bibinfo{pages}{71078--71094}.
\newblock


\bibitem[Maji et~al\mbox{.}(2013)]%
        {zs_cls9}
\bibfield{author}{\bibinfo{person}{Subhransu Maji}, \bibinfo{person}{Esa Rahtu}, \bibinfo{person}{Juho Kannala}, \bibinfo{person}{Matthew Blaschko}, {and} \bibinfo{person}{Andrea Vedaldi}.} \bibinfo{year}{2013}\natexlab{}.
\newblock \showarticletitle{Fine-grained visual classification of aircraft}.
\newblock \bibinfo{journal}{\emph{arXiv preprint arXiv:1306.5151}} (\bibinfo{year}{2013}).
\newblock


\bibitem[Ma{\~n}as et~al\mbox{.}(2022)]%
        {manas2022mapl}
\bibfield{author}{\bibinfo{person}{Oscar Ma{\~n}as}, \bibinfo{person}{Pau Rodriguez}, \bibinfo{person}{Saba Ahmadi}, \bibinfo{person}{Aida Nematzadeh}, \bibinfo{person}{Yash Goyal}, {and} \bibinfo{person}{Aishwarya Agrawal}.} \bibinfo{year}{2022}\natexlab{}.
\newblock \showarticletitle{Mapl: Parameter-efficient adaptation of unimodal pre-trained models for vision-language few-shot prompting}.
\newblock \bibinfo{journal}{\emph{arXiv preprint arXiv:2210.07179}} (\bibinfo{year}{2022}).
\newblock


\bibitem[Minderer et~al\mbox{.}(2022)]%
        {minderer2022simple}
\bibfield{author}{\bibinfo{person}{Matthias Minderer}, \bibinfo{person}{Alexey Gritsenko}, \bibinfo{person}{Austin Stone}, \bibinfo{person}{Maxim Neumann}, \bibinfo{person}{Dirk Weissenborn}, \bibinfo{person}{Alexey Dosovitskiy}, \bibinfo{person}{Aravindh Mahendran}, \bibinfo{person}{Anurag Arnab}, \bibinfo{person}{Mostafa Dehghani}, \bibinfo{person}{Zhuoran Shen}, {et~al\mbox{.}}} \bibinfo{year}{2022}\natexlab{}.
\newblock \showarticletitle{Simple open-vocabulary object detection}. In \bibinfo{booktitle}{\emph{ECCV}}. Springer, \bibinfo{pages}{728--755}.
\newblock


\bibitem[Morris et~al\mbox{.}(2020)]%
        {Grammar}
\bibfield{author}{\bibinfo{person}{John~X Morris}, \bibinfo{person}{Eli Lifland}, \bibinfo{person}{Jin~Yong Yoo}, \bibinfo{person}{Jake Grigsby}, \bibinfo{person}{Di Jin}, {and} \bibinfo{person}{Yanjun Qi}.} \bibinfo{year}{2020}\natexlab{}.
\newblock \showarticletitle{Textattack: A framework for adversarial attacks, data augmentation, and adversarial training in nlp}.
\newblock \bibinfo{journal}{\emph{arXiv preprint arXiv:2005.05909}} (\bibinfo{year}{2020}).
\newblock


\bibitem[Mu et~al\mbox{.}(2022)]%
        {mu2022slip}
\bibfield{author}{\bibinfo{person}{Norman Mu}, \bibinfo{person}{Alexander Kirillov}, \bibinfo{person}{David Wagner}, {and} \bibinfo{person}{Saining Xie}.} \bibinfo{year}{2022}\natexlab{}.
\newblock \showarticletitle{Slip: Self-supervision meets language-image pre-training}. In \bibinfo{booktitle}{\emph{ECCV}}. Springer, \bibinfo{pages}{529--544}.
\newblock


\bibitem[Nilsback and Zisserman(2008)]%
        {zs_cls4}
\bibfield{author}{\bibinfo{person}{Maria-Elena Nilsback} {and} \bibinfo{person}{Andrew Zisserman}.} \bibinfo{year}{2008}\natexlab{}.
\newblock \showarticletitle{Automated flower classification over a large number of classes}. In \bibinfo{booktitle}{\emph{2008 Sixth Indian conference on computer vision, graphics \& image processing}}. IEEE, \bibinfo{pages}{722--729}.
\newblock


\bibitem[Oord et~al\mbox{.}(2018a)]%
        {infonce}
\bibfield{author}{\bibinfo{person}{Aaron van~den Oord}, \bibinfo{person}{Yazhe Li}, {and} \bibinfo{person}{Oriol Vinyals}.} \bibinfo{year}{2018}\natexlab{a}.
\newblock \showarticletitle{Representation learning with contrastive predictive coding}.
\newblock \bibinfo{journal}{\emph{arXiv preprint arXiv:1807.03748}} (\bibinfo{year}{2018}).
\newblock


\bibitem[Oord et~al\mbox{.}(2018b)]%
        {oord2018representation}
\bibfield{author}{\bibinfo{person}{Aaron van~den Oord}, \bibinfo{person}{Yazhe Li}, {and} \bibinfo{person}{Oriol Vinyals}.} \bibinfo{year}{2018}\natexlab{b}.
\newblock \showarticletitle{Representation learning with contrastive predictive coding}.
\newblock \bibinfo{journal}{\emph{arXiv preprint arXiv:1807.03748}} (\bibinfo{year}{2018}).
\newblock


\bibitem[Parcalabescu et~al\mbox{.}(2021)]%
        {parcalabescu2021valse}
\bibfield{author}{\bibinfo{person}{Letitia Parcalabescu}, \bibinfo{person}{Michele Cafagna}, \bibinfo{person}{Lilitta Muradjan}, \bibinfo{person}{Anette Frank}, \bibinfo{person}{Iacer Calixto}, {and} \bibinfo{person}{Albert Gatt}.} \bibinfo{year}{2021}\natexlab{}.
\newblock \showarticletitle{VALSE: A task-independent benchmark for vision and language models centered on linguistic phenomena}.
\newblock \bibinfo{journal}{\emph{arXiv preprint arXiv:2112.07566}} (\bibinfo{year}{2021}).
\newblock


\bibitem[Parkhi et~al\mbox{.}(2012)]%
        {zs_cls3}
\bibfield{author}{\bibinfo{person}{Omkar~M Parkhi}, \bibinfo{person}{Andrea Vedaldi}, \bibinfo{person}{Andrew Zisserman}, {and} \bibinfo{person}{CV Jawahar}.} \bibinfo{year}{2012}\natexlab{}.
\newblock \showarticletitle{Cats and dogs}. In \bibinfo{booktitle}{\emph{CVPR}}. IEEE, \bibinfo{pages}{3498--3505}.
\newblock


\bibitem[Peng et~al\mbox{.}(2024)]%
        {peng2024synthesize}
\bibfield{author}{\bibinfo{person}{Wujian Peng}, \bibinfo{person}{Sicheng Xie}, \bibinfo{person}{Zuyao You}, \bibinfo{person}{Shiyi Lan}, {and} \bibinfo{person}{Zuxuan Wu}.} \bibinfo{year}{2024}\natexlab{}.
\newblock \showarticletitle{Synthesize Diagnose and Optimize: Towards Fine-Grained Vision-Language Understanding}. In \bibinfo{booktitle}{\emph{CVPR}}. \bibinfo{pages}{13279--13288}.
\newblock


\bibitem[Plummer et~al\mbox{.}(2015)]%
        {zs_ret_flcker}
\bibfield{author}{\bibinfo{person}{Bryan~A Plummer}, \bibinfo{person}{Liwei Wang}, \bibinfo{person}{Chris~M Cervantes}, \bibinfo{person}{Juan~C Caicedo}, \bibinfo{person}{Julia Hockenmaier}, {and} \bibinfo{person}{Svetlana Lazebnik}.} \bibinfo{year}{2015}\natexlab{}.
\newblock \showarticletitle{Flickr30k entities: Collecting region-to-phrase correspondences for richer image-to-sentence models}. In \bibinfo{booktitle}{\emph{ICCV}}. \bibinfo{pages}{2641--2649}.
\newblock


\bibitem[Radford et~al\mbox{.}(2021a)]%
        {clip}
\bibfield{author}{\bibinfo{person}{Alec Radford}, \bibinfo{person}{Jong~Wook Kim}, \bibinfo{person}{Chris Hallacy}, \bibinfo{person}{Aditya Ramesh}, \bibinfo{person}{Gabriel Goh}, \bibinfo{person}{Sandhini Agarwal}, \bibinfo{person}{Girish Sastry}, \bibinfo{person}{Amanda Askell}, \bibinfo{person}{Pamela Mishkin}, \bibinfo{person}{Jack Clark}, {et~al\mbox{.}}} \bibinfo{year}{2021}\natexlab{a}.
\newblock \showarticletitle{Learning transferable visual models from natural language supervision}. In \bibinfo{booktitle}{\emph{ICML}}. PmLR, \bibinfo{pages}{8748--8763}.
\newblock


\bibitem[Radford et~al\mbox{.}(2021b)]%
        {radford2021learning}
\bibfield{author}{\bibinfo{person}{Alec Radford}, \bibinfo{person}{Jong~Wook Kim}, \bibinfo{person}{Chris Hallacy}, \bibinfo{person}{Aditya Ramesh}, \bibinfo{person}{Gabriel Goh}, \bibinfo{person}{Sandhini Agarwal}, \bibinfo{person}{Girish Sastry}, \bibinfo{person}{Amanda Askell}, \bibinfo{person}{Pamela Mishkin}, \bibinfo{person}{Jack Clark}, {et~al\mbox{.}}} \bibinfo{year}{2021}\natexlab{b}.
\newblock \showarticletitle{Learning transferable visual models from natural language supervision}. In \bibinfo{booktitle}{\emph{ICML}}. PMLR, \bibinfo{pages}{8748--8763}.
\newblock


\bibitem[Thrush et~al\mbox{.}(2022)]%
        {thrush2022winoground}
\bibfield{author}{\bibinfo{person}{Tristan Thrush}, \bibinfo{person}{Ryan Jiang}, \bibinfo{person}{Max Bartolo}, \bibinfo{person}{Amanpreet Singh}, \bibinfo{person}{Adina Williams}, \bibinfo{person}{Douwe Kiela}, {and} \bibinfo{person}{Candace Ross}.} \bibinfo{year}{2022}\natexlab{}.
\newblock \showarticletitle{Winoground: Probing vision and language models for visio-linguistic compositionality}. In \bibinfo{booktitle}{\emph{CVPR}}. \bibinfo{pages}{5238--5248}.
\newblock


\bibitem[Wang et~al\mbox{.}(2022)]%
        {wang2022image}
\bibfield{author}{\bibinfo{person}{Wenhui Wang}, \bibinfo{person}{Hangbo Bao}, \bibinfo{person}{Li Dong}, \bibinfo{person}{Johan Bjorck}, \bibinfo{person}{Zhiliang Peng}, \bibinfo{person}{Qiang Liu}, \bibinfo{person}{Kriti Aggarwal}, \bibinfo{person}{Owais~Khan Mohammed}, \bibinfo{person}{Saksham Singhal}, \bibinfo{person}{Subhojit Som}, {et~al\mbox{.}}} \bibinfo{year}{2022}\natexlab{}.
\newblock \showarticletitle{Image as a foreign language: Beit pretraining for all vision and vision-language tasks}.
\newblock \bibinfo{journal}{\emph{arXiv preprint arXiv:2208.10442}} (\bibinfo{year}{2022}).
\newblock


\bibitem[Wu et~al\mbox{.}(2023)]%
        {wu2023tinyclip}
\bibfield{author}{\bibinfo{person}{Kan Wu}, \bibinfo{person}{Houwen Peng}, \bibinfo{person}{Zhenghong Zhou}, \bibinfo{person}{Bin Xiao}, \bibinfo{person}{Mengchen Liu}, \bibinfo{person}{Lu Yuan}, \bibinfo{person}{Hong Xuan}, \bibinfo{person}{Michael Valenzuela}, \bibinfo{person}{Xi~Stephen Chen}, \bibinfo{person}{Xinggang Wang}, {et~al\mbox{.}}} \bibinfo{year}{2023}\natexlab{}.
\newblock \showarticletitle{Tinyclip: Clip distillation via affinity mimicking and weight inheritance}. In \bibinfo{booktitle}{\emph{ICCV}}. \bibinfo{pages}{21970--21980}.
\newblock


\bibitem[Wu et~al\mbox{.}(2024b)]%
        {object_3}
\bibfield{author}{\bibinfo{person}{Size Wu}, \bibinfo{person}{Wenwei Zhang}, \bibinfo{person}{Lumin Xu}, \bibinfo{person}{Sheng Jin}, \bibinfo{person}{Xiangtai Li}, \bibinfo{person}{Wentao Liu}, {and} \bibinfo{person}{Chen~Change Loy}.} \bibinfo{year}{2024}\natexlab{b}.
\newblock \showarticletitle{{CLIPS}elf: Vision Transformer Distills Itself for Open-Vocabulary Dense Prediction}. In \bibinfo{booktitle}{\emph{The Twelfth International Conference on Learning Representations}}.
\newblock
\urldef\tempurl%
\url{https://openreview.net/forum?id=DjzvJCRsVf}
\showURL{%
\tempurl}


\bibitem[Wu et~al\mbox{.}(2024a)]%
        {wu2024clip2uda}
\bibfield{author}{\bibinfo{person}{Yao Wu}, \bibinfo{person}{Mingwei Xing}, \bibinfo{person}{Yachao Zhang}, \bibinfo{person}{Yuan Xie}, {and} \bibinfo{person}{Yanyun Qu}.} \bibinfo{year}{2024}\natexlab{a}.
\newblock \showarticletitle{Clip2uda: Making frozen clip reward unsupervised domain adaptation in 3d semantic segmentation}. In \bibinfo{booktitle}{\emph{ACM MM}}. \bibinfo{pages}{8662--8671}.
\newblock


\bibitem[Xiao et~al\mbox{.}(2010)]%
        {zs_cls5}
\bibfield{author}{\bibinfo{person}{Jianxiong Xiao}, \bibinfo{person}{James Hays}, \bibinfo{person}{Krista~A Ehinger}, \bibinfo{person}{Aude Oliva}, {and} \bibinfo{person}{Antonio Torralba}.} \bibinfo{year}{2010}\natexlab{}.
\newblock \showarticletitle{Sun database: Large-scale scene recognition from abbey to zoo}. In \bibinfo{booktitle}{\emph{CVPR}}. IEEE, \bibinfo{pages}{3485--3492}.
\newblock


\bibitem[Yang et~al\mbox{.}(2023)]%
        {yang2023alip}
\bibfield{author}{\bibinfo{person}{Kaicheng Yang}, \bibinfo{person}{Jiankang Deng}, \bibinfo{person}{Xiang An}, \bibinfo{person}{Jiawei Li}, \bibinfo{person}{Ziyong Feng}, \bibinfo{person}{Jia Guo}, \bibinfo{person}{Jing Yang}, {and} \bibinfo{person}{Tongliang Liu}.} \bibinfo{year}{2023}\natexlab{}.
\newblock \showarticletitle{Alip: Adaptive language-image pre-training with synthetic caption}. In \bibinfo{booktitle}{\emph{ICCV}}. \bibinfo{pages}{2922--2931}.
\newblock


\bibitem[Yang et~al\mbox{.}(2024)]%
        {yang2024clip}
\bibfield{author}{\bibinfo{person}{Kaicheng Yang}, \bibinfo{person}{Tiancheng Gu}, \bibinfo{person}{Xiang An}, \bibinfo{person}{Haiqiang Jiang}, \bibinfo{person}{Xiangzi Dai}, \bibinfo{person}{Ziyong Feng}, \bibinfo{person}{Weidong Cai}, {and} \bibinfo{person}{Jiankang Deng}.} \bibinfo{year}{2024}\natexlab{}.
\newblock \showarticletitle{Clip-cid: Efficient clip distillation via cluster-instance discrimination}.
\newblock \bibinfo{journal}{\emph{AAAI}} (\bibinfo{year}{2024}).
\newblock


\bibitem[Yao et~al\mbox{.}(2021)]%
        {yao2021filip}
\bibfield{author}{\bibinfo{person}{Lewei Yao}, \bibinfo{person}{Runhui Huang}, \bibinfo{person}{Lu Hou}, \bibinfo{person}{Guansong Lu}, \bibinfo{person}{Minzhe Niu}, \bibinfo{person}{Hang Xu}, \bibinfo{person}{Xiaodan Liang}, \bibinfo{person}{Zhenguo Li}, \bibinfo{person}{Xin Jiang}, {and} \bibinfo{person}{Chunjing Xu}.} \bibinfo{year}{2021}\natexlab{}.
\newblock \showarticletitle{Filip: Fine-grained interactive language-image pre-training}.
\newblock \bibinfo{journal}{\emph{arXiv preprint arXiv:2111.07783}} (\bibinfo{year}{2021}).
\newblock


\bibitem[Yu et~al\mbox{.}(2023)]%
        {yu2023convolutions}
\bibfield{author}{\bibinfo{person}{Qihang Yu}, \bibinfo{person}{Ju He}, \bibinfo{person}{Xueqing Deng}, \bibinfo{person}{Xiaohui Shen}, {and} \bibinfo{person}{Liang-Chieh Chen}.} \bibinfo{year}{2023}\natexlab{}.
\newblock \showarticletitle{Convolutions die hard: Open-vocabulary segmentation with single frozen convolutional clip}.
\newblock \bibinfo{journal}{\emph{NIPS}}  \bibinfo{volume}{36} (\bibinfo{year}{2023}), \bibinfo{pages}{32215--32234}.
\newblock


\bibitem[Yuksekgonul et~al\mbox{.}(2022)]%
        {yuksekgonul2022and}
\bibfield{author}{\bibinfo{person}{Mert Yuksekgonul}, \bibinfo{person}{Federico Bianchi}, \bibinfo{person}{Pratyusha Kalluri}, \bibinfo{person}{Dan Jurafsky}, {and} \bibinfo{person}{James Zou}.} \bibinfo{year}{2022}\natexlab{}.
\newblock \showarticletitle{When and why vision-language models behave like bags-of-words, and what to do about it?}
\newblock \bibinfo{journal}{\emph{arXiv preprint arXiv:2210.01936}} (\bibinfo{year}{2022}).
\newblock


\bibitem[Zang et~al\mbox{.}(2022)]%
        {zang2022open}
\bibfield{author}{\bibinfo{person}{Yuhang Zang}, \bibinfo{person}{Wei Li}, \bibinfo{person}{Kaiyang Zhou}, \bibinfo{person}{Chen Huang}, {and} \bibinfo{person}{Chen~Change Loy}.} \bibinfo{year}{2022}\natexlab{}.
\newblock \showarticletitle{Open-vocabulary detr with conditional matching}. In \bibinfo{booktitle}{\emph{ECCV}}. Springer, \bibinfo{pages}{106--122}.
\newblock


\bibitem[Zeng et~al\mbox{.}(2021a)]%
        {xvlm}
\bibfield{author}{\bibinfo{person}{Yan Zeng}, \bibinfo{person}{Xinsong Zhang}, {and} \bibinfo{person}{Hang Li}.} \bibinfo{year}{2021}\natexlab{a}.
\newblock \showarticletitle{Multi-Grained Vision Language Pre-Training: Aligning Texts with Visual Concepts}.
\newblock \bibinfo{journal}{\emph{arXiv preprint arXiv:2111.08276}} (\bibinfo{year}{2021}).
\newblock


\bibitem[Zeng et~al\mbox{.}(2021b)]%
        {zeng2021multi}
\bibfield{author}{\bibinfo{person}{Yan Zeng}, \bibinfo{person}{Xinsong Zhang}, {and} \bibinfo{person}{Hang Li}.} \bibinfo{year}{2021}\natexlab{b}.
\newblock \showarticletitle{Multi-grained vision language pre-training: Aligning texts with visual concepts}.
\newblock \bibinfo{journal}{\emph{arXiv preprint arXiv:2111.08276}} (\bibinfo{year}{2021}).
\newblock


\bibitem[Zhai et~al\mbox{.}(2023)]%
        {siglip}
\bibfield{author}{\bibinfo{person}{Xiaohua Zhai}, \bibinfo{person}{Basil Mustafa}, \bibinfo{person}{Alexander Kolesnikov}, {and} \bibinfo{person}{Lucas Beyer}.} \bibinfo{year}{2023}\natexlab{}.
\newblock \showarticletitle{Sigmoid loss for language image pre-training}. In \bibinfo{booktitle}{\emph{ICCV}}. \bibinfo{pages}{11975--11986}.
\newblock


\bibitem[Zhang et~al\mbox{.}(2024a)]%
        {zhang2024contrasting}
\bibfield{author}{\bibinfo{person}{Le Zhang}, \bibinfo{person}{Rabiul Awal}, {and} \bibinfo{person}{Aishwarya Agrawal}.} \bibinfo{year}{2024}\natexlab{a}.
\newblock \showarticletitle{Contrasting Intra-Modal and Ranking Cross-Modal Hard Negatives to Enhance Visio-Linguistic Compositional Understanding}. In \bibinfo{booktitle}{\emph{CVPR}}. \bibinfo{pages}{13774--13784}.
\newblock


\bibitem[Zhang et~al\mbox{.}(2024b)]%
        {zhang2024alignclip}
\bibfield{author}{\bibinfo{person}{Lu Zhang}, \bibinfo{person}{Ke Yan}, {and} \bibinfo{person}{Shouhong Ding}.} \bibinfo{year}{2024}\natexlab{b}.
\newblock \showarticletitle{AlignCLIP: Align Multi Domains of Texts Input for CLIP models with Object-IoU Loss}. In \bibinfo{booktitle}{\emph{ACM MM}}. \bibinfo{pages}{1092--1100}.
\newblock


\bibitem[Zhu et~al\mbox{.}(2023)]%
        {zhu2023minigpt}
\bibfield{author}{\bibinfo{person}{Deyao Zhu}, \bibinfo{person}{Jun Chen}, \bibinfo{person}{Xiaoqian Shen}, \bibinfo{person}{Xiang Li}, {and} \bibinfo{person}{Mohamed Elhoseiny}.} \bibinfo{year}{2023}\natexlab{}.
\newblock \showarticletitle{Minigpt-4: Enhancing vision-language understanding with advanced large language models}.
\newblock \bibinfo{journal}{\emph{arXiv preprint arXiv:2304.10592}} (\bibinfo{year}{2023}).
\newblock


\end{thebibliography}
\clearpage
\appendix
\section{Supplementary Material}
\subsection{Training Details} 
Details of the hyperparameter configurations are presented in Table~\ref{tab:hyper_parameter}. An asterisk (*) indicates that the batch size for our positive image-text pairs is 256, and each positive text is paired with four negative texts during training.

\begin{table}[h!]
    \centering
    \begin{tabular}{cc}
    \toprule
    \multicolumn{2}{c}{\textbf{Hyperparameters}} \\
    \midrule
    Batch size & 256* \\
    Optimizer & AdamW\\
    Weight decay & 0.1\\
    Adam $\beta$ & (0.9,0.98)\\
    Adam $\epsilon$ & 1e-6\\
    Learning rate & 1e-6 \\
    Learning rate schedule & cosine decay\\
    Ema $\alpha$ & 0.9996\\
    $(\lambda_1,\lambda_2,\lambda_3)$ & (0.1,0.1,0.005)\\
    Epochs & 5 \\
    Training GPUs & $8\times$V100\\
    \bottomrule
    \end{tabular}
    \caption{Detailed hyper-parameters for training DeGLA.}
    \label{tab:hyper_parameter}
    \vspace{-7mm}
\end{table}

\begin{table}[t!]
\centering
\resizebox{\linewidth}{!}{
\begin{tabular}{lcccr}
\toprule
\multicolumn{1}{l}{Dataset} & \multicolumn{1}{c}{Classes} & \multicolumn{1}{c}{Train size} & \multicolumn{1}{c}{Test size} & \multicolumn{1}{c}{Evaluation metric} \\
\midrule
Food101                        & 102                             & 75,750                             & 25,250                            & accuracy                                  \\
CIFAR10                        & 10                              & 50,000                             & 10,000                            & accuracy                                  \\
CIFAR100                       & 100                             & 50,000                             & 10,000                            & accuracy                                  \\
SUN397                          & 397                             & 19,850                             & 19,850                            & accuracy                                  \\
Cars                   & 196                             & 8,144                              & 8,041                             & accuracy                                  \\
Aircraft                   & 100                             & 6,667                              & 3,333                             & mean per class                            \\
DTD           & 47                              & 3,760                              & 1,880                             & accuracy                                  \\
Pets                & 37                              & 3,680                              & 3,669                             & mean per class                            \\
Caltech101                     & 101                             & 3,000                              & 5,677                             & mean-per-class                            \\
Flowers                  & 102                             & 2,040                              & 6,149                             & mean per class                            \\

ImageNet                        & 1000                            & 1,281,167                          & 50,000                            & accuracy                                  \\
\bottomrule
\end{tabular}}
\caption{List of linear probe datasets with the data distribution and evaluation metrics.}
\vspace{-7mm}
\label{tab:linearprobe_datasets}
\end{table}

\begin{table}[t!]
\centering
\resizebox{\linewidth}{!}{
\begin{tabular}{lccc}
\toprule
\textbf{Dataset} & \textbf{Test Images} & \textbf{Captions per Image} & \textbf{Evaluation Protocol} \\
\midrule
MSCOCO & 5,000 & 5 & Image-to-Text \& Text-to-Image \\
Flickr30k & 1,000 & 5 & Image-to-Text \& Text-to-Image \\
\bottomrule
\end{tabular}
}
\caption{Zero-shot image-text retrieval evaluation settings.}
\label{tab:retrieval_settings}\
\vspace{-7mm}
\end{table}

\begin{figure*}[!t]
    \centering
    \includegraphics[width=\linewidth]{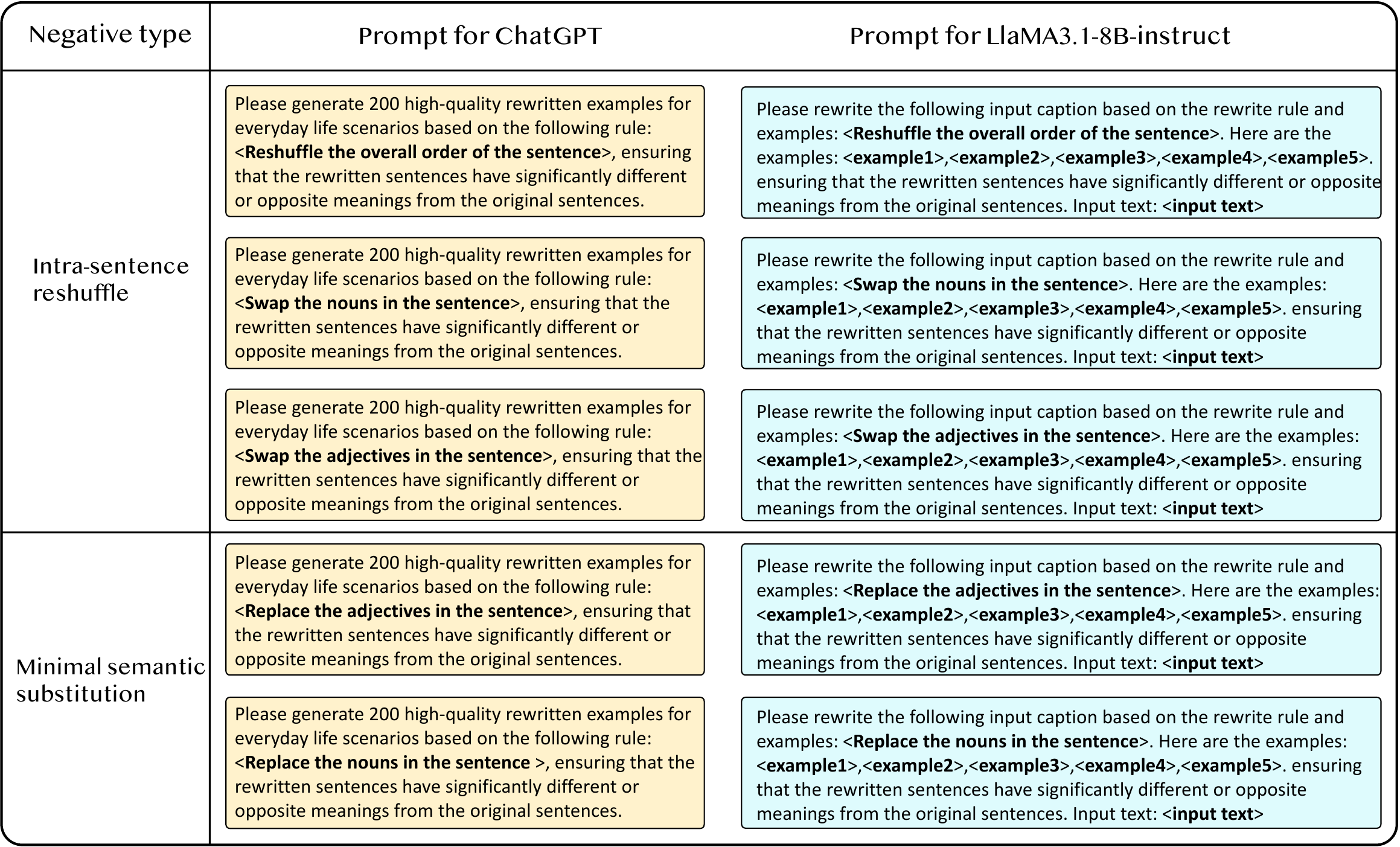}
    \vspace{-7mm}
    \caption{Details of the prompt utilized for generating high-quality examples and negative captions}
    \label{fig:prompt}
    \vspace{-2mm}
    
\end{figure*}

\begin{table*}[t!]
    \centering
    \begin{tabular}{ccc}
    \toprule
       Benchmark & \#Examples & Composionablity \\
       \midrule
       ARO~\cite{yuksekgonul2022and} & 82695 & Relation,Attribution,Order\\
       VALSE~\cite{parcalabescu2021valse} & 8309 & Existence, Plurality, Counting, Spatial Relations, Actions,Entity Coreference\\
       SugarCrepe~\cite{hsieh2024sugarcrepe} & 5948 & Relation,Attribution,Order,Semantic Substitution\\
        \bottomrule
    \end{tabular}
    \caption{Details of compositional reasoning benchmarks.}
    \label{tab:com_bench_details}
    \vspace{-5mm}
\end{table*}
\begin{table*}[h]
    \centering
    \resizebox{1.0\textwidth}{!}{
    \begin{tabular}{ccccccccc}
    \toprule
    \multirow{2}{*}{Method} & \multirow{2}{*}{Pretrain} & \multirow{2}{*}{Finetune} & \multirow{2}{*}{Hard negative data} & \multirow{2}{*}{Finegrained loss} & \multicolumn{3}{c}{Hard negative type} & \multirow{2}{*}{Self distillation} \\
    \cmidrule(lr){6-8}
    & & & & & Rule-based & Un-masking based & LLM-based & \\
    \midrule
    CLIP~\cite{clip}                     & \ding{51} &  &  &  &  &  &  &  \\
    \midrule
    NegCLIP~\cite{yuksekgonul2022and}   &  & \ding{51} & \ding{51} &  &\ding{51}  &  &  &  \\
    Structure-CLIP~\cite{huang2024structure} &  & \ding{51} & \ding{51} &  & \ding{51} &  &  &  \\
    CE-CLIP~\cite{zhang2024contrasting} &  & \ding{51} & \ding{51} & \ding{51} & \ding{51} & \ding{51} &  &  \\
    DeGLA (ours)                        &  & \ding{51} & \ding{51} & \ding{51} & & & \ding{51} & \ding{51} \\
    \bottomrule
    \end{tabular}
    }
    \caption{Comparison of different methods.}
    \label{tab:comparision_framework}
    \vspace{-7mm}
\end{table*}

\subsection{Detatils of General Understanding Downstream Datasets}
\noindent\textbf{Zero-shot Classification.} 
Following the previous work RWKV-CLIP~\cite{rwkvclip}, we evaluate the zero-shot classification performance of the models on 11 datasets. The prompt used in zero-shot classification is presented in Table~\ref{tab:prompt}.

\noindent\textbf{Linear Probe.} 
The datasets used for the linear probe evaluation are the same as those used for zero-shot classification. Details on each dataset and the corresponding evaluation metrics are provided in Table~\ref{tab:linearprobe_datasets}.

\noindent\textbf{Zero-shot Image-Text Retrieval.} 
The detail of zero-shot image-text retrieval benchmark is presented in Table~\ref{tab:retrieval_settings}.
We evaluate retrieval performance on MS-COCO and Flickr30k following standard protocols. For MS-COCO, we use the 1K test set from ~\cite{karpathy2015deep}, while Flickr30k employs 1,000 images with 5 captions each. In both datasets, we assess bidirectional retrieval (image-to-text and text-to-image) using pre-trained model embeddings without fine-tuning. Performance is measured by Recall@K (R@1, R@5, R@10), representing the percentage of correct matches in the top-K results.

\subsection{Details of Compositional Reasoning Benchmarks}
In Table~\ref{tab:com_bench_details}, we summarize an overview of the three compositional reasoning benchmarks employed in this work.

\noindent\textbf{ARO.}
ARO~\cite{yuksekgonul2022and} benchmark is used to probe the VLM's understanding of relations, attributes, and order in image-text data. Specifically, the Relation task involves swapping objects in the text, the Attribute task involves switching attributes in the text, and the Order task involves disrupting the sequence of the entire sentence.
\begin{figure*}[!t]
    \centering
    \includegraphics[width=0.9\linewidth]{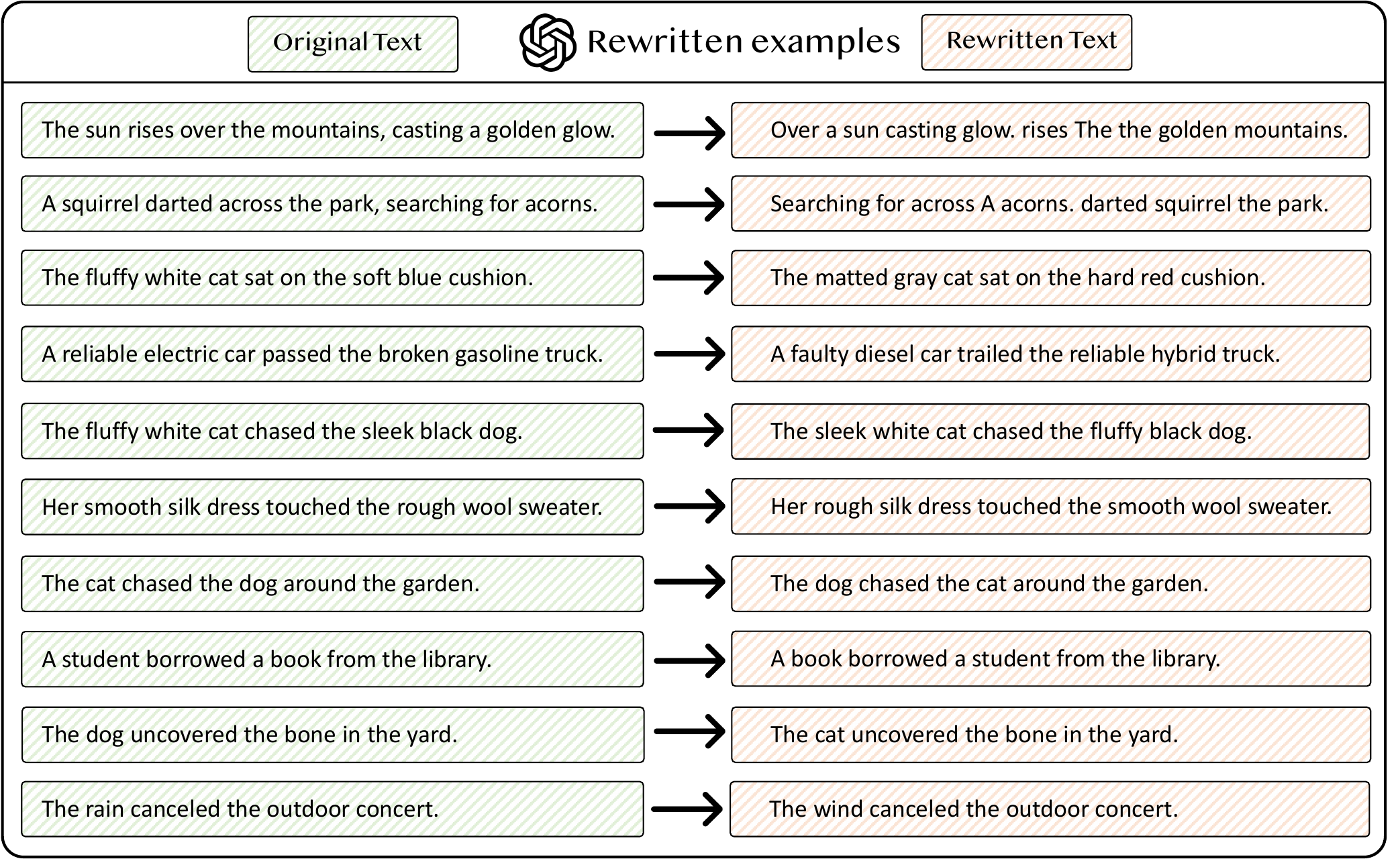}
    \vspace{-3mm}
    \caption{Rewritten examples generated by ChatGPT.}
    \Description{A figure showing rewritten text examples generated by ChatGPT. Each example consists of an input sentence and its rewritten version.}
    \label{fig:rewritten_example}
    \vspace{-3mm}
\end{figure*}
\begin{figure*}[!t]
    \centering
    \includegraphics[width=0.9\linewidth]{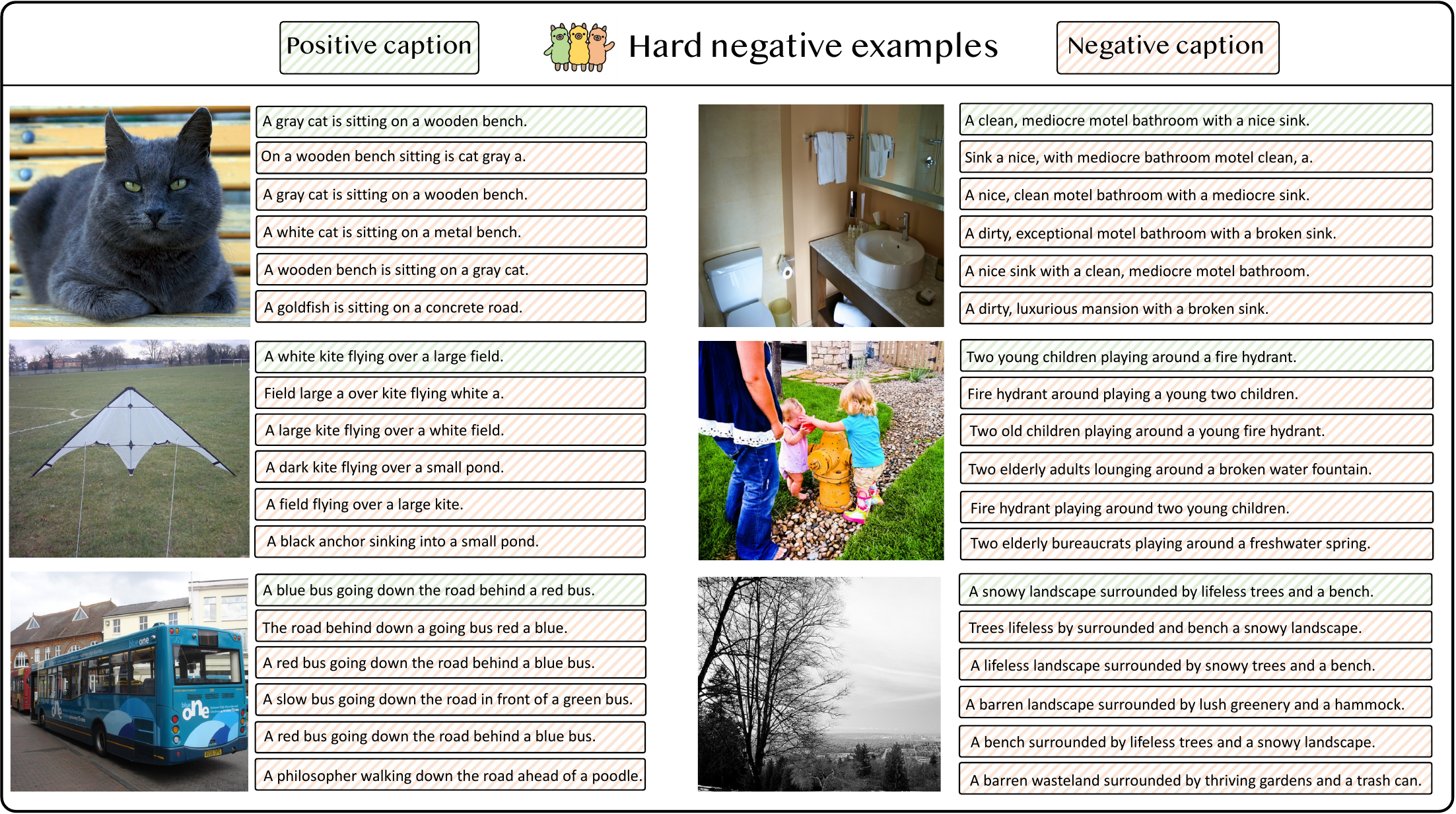}
    \vspace{-3mm}
    \caption{Hard negative examples generated by LLaMA3.1-8B-instruct.}
    \label{fig:example}
    \vspace{-3mm}
\end{figure*}

\noindent\textbf{VALSE.}
The VASLE~\cite{parcalabescu2021valse} benchmark assesses the compositional understanding of Visual Language Models (VLMs) across six dimensions: existence, plurality, counting, spatial relations, actions, and entity coreference. The existence subset employs a single cue to evaluate whether models can discern the presence or absence of specific entities in images. The plurality subset similarly uses a single cue to determine if models can differentiate between singular and plural noun phrases depicted in images, such as ``exactly one flower'' vs. ``some flowers''. The counting subset, which includes balanced, adversarial, and small-number scenarios, tests the models' ability to confirm the accuracy of a stated number of entities against those shown in the image. The spatial relations subset uses a single cue to assess model performance in identifying different spatial relationships, creating foils by modifying the spatial preposition in the original caption. The actions subset, encompassing action replacement and actant swap settings, evaluates the models' capability to (i) match the described action with the depicted action, and (ii) accurately identify the agents performing and receiving the action. Finally, the coreference subset examines the models' proficiency in resolving pronominal references within multimodal contexts, covering both pronouns linked to noun phrases visually grounded in the image and deictic or image-level references.

\begin{table*}[h!]
\centering
\resizebox{\linewidth}{!}{
\begin{tabular}{llll}
\toprule
\multicolumn{4}{l}{\bf CIFAR 10 \& CIFAR 100} \\
a photo of a \{label\}. &
a blurry photo of a \{label\}. &
a black and white photo of a \{label\}. &
a low contrast photo of a \{label\}. \\
a high contrast photo of a \{label\}. &
a bad photo of a \{label\}. &
a good photo of a \{label\}. &
a photo of a small \{label\}. \\
a photo of a big \{label\}.&
a photo of the \{label\}.&
a blurry photo of the \{label\}.&
a black and white photo of the \{label\}. \\
a low contrast photo of the \{label\}.&
a high contrast photo of the \{label\}.&
a bad photo of the \{label\}.&
a good photo of the \{label\}. \\
a photo of the small \{label\}.&
a photo of the big \{label\}.& & \\
\midrule
\multicolumn{4}{l}{\bf Food101} \\
a photo of \{label\}, a type of food. & & \\
\midrule
\multicolumn{4}{l}{\bf Caltech101} \\
a photo of a \{label\}. &
a painting of a \{label\}. &
a plastic \{label\}. &
a sculpture of a \{label\}. \\
a sketch of a \{label\}. &
a tattoo of a \{label\}. &
a toy \{label\}. &
a rendition of a \{label\}. \\
a embroidered \{label\}. &
a cartoon \{label\}. &
a \{label\} in a video game. &
a plushie \{label\}. \\
an origami \{label\}. &
art of a \{label\}. &
graffiti of a \{label\}. &
a drawing of a \{label\}. \\
a doodle of a \{label\}. &
a photo of the \{label\}. &
a painting of the \{label\}.&
the plastic \{label\}. \\
a sculpture of the \{label\}.&
a sketch of the \{label\}.&
a tattoo of the \{label\}.&
the toy \{label\}. \\
a rendition of the \{label\}.&
the embroidered \{label\}.&
the cartoon \{label\}.&
the \{label\} in a video game. \\
the plushie \{label\}.&
the origami \{label\}.&
art of the \{label\}.&
graffiti of the \{label\}. \\
a drawing of the \{label\}.&
a doodle of the \{label\}.& & \\
\midrule
\multicolumn{4}{l}{\bf Stanford Cars} \\
a photo of a \{label\}.&
a photo of the \{label\}.&
a photo of my \{label\}.&
i love my \{label\}! \\
a photo of my dirty \{label\}.&
a photo of my clean \{label\}.&
a photo of my new \{label\}.&
a photo of my old \{label\}. \\
\midrule
\multicolumn{4}{l}{\bf DTD} \\
a photo of a \{label\} texture.&
a photo of a \{label\} pattern.&
a photo of a \{label\} thing.&
a photo of a \{label\} object. \\
a photo of the \{label\} texture. &
a photo of the \{label\} pattern. &
a photo of the \{label\} thing. &
a photo of the \{label\} object. \\
\midrule
\multicolumn{4}{l}{\bf FGVC Aircraft} \\
a photo of a \{label\}, a type of aircraft.&
a photo of the \{label\}, a type of aircraft.& & \\
\midrule
\multicolumn{4}{l}{\bf Flowers102} \\
a photo of a \{label\}, a type of flower. &&& \\
\midrule
\multicolumn{4}{l}{\bf Pets } \\
a photo of a \{label\}, a type of pet.&&& \\
\midrule
\multicolumn{4}{l}{\bf  SUN39} \\
a photo of a \{label\}.&
a photo of the \{label\}.&& \\
\midrule
\multicolumn{4}{l}{\bf  ImageNet} \\
a bad photo of a \{label\}. & 
a photo of many \{label\}. &
a sculpture of a \{label\}. &
a photo of the hard to see \{label\}. \\
a low resolution photo of the \{label\}. & 
a rendering of a \{label\}. &
graffiti of a \{label\}. &
a bad photo of the \{label\}.  \\
a cropped photo of the \{label\}. &
a tattoo of a \{label\}. & 
the embroidered \{label\}. &
a photo of a hard to see \{label\}.  \\
a bright photo of a \{label\}.&
a photo of a clean \{label\}.&
a photo of a dirty \{label\}.&
a dark photo of the \{label\}. \\
a drawing of a \{label\}.&
a photo of my \{label\}.&
the plastic \{label\}.&
a photo of the cool \{label\}. \\
a close-up photo of a \{label\}.&
a black and white photo of the \{label\}.&
a painting of the \{label\}.&
a painting of a \{label\}. \\
a pixelated photo of the \{label\}.& 
a sculpture of the \{label\}.&
a bright photo of the \{label\}.&
a cropped photo of a \{label\}. \\
a plastic \{label\}.&
a photo of the dirty \{label\}.& 
a jpeg corrupted photo of a \{label\}.&
a blurry photo of the \{label\}. \\
a photo of the \{label\}.&
a good photo of the \{label\}.&
a rendering of the \{label\}.&
a \{label\} in a video game. \\
a photo of one \{label\}.&
a doodle of a \{label\}.&
a close-up photo of the \{label\}.&
a photo of a \{label\}. \\
the origami \{label\}.&
the \{label\} in a video game.&
a sketch of a \{label\}.&
a doodle of the \{label\}. \\
an origami \{label\}.&
a low resolution photo of a \{label\}.&
the toy \{label\}.&
a rendition of the \{label\}. \\
a photo of the clean \{label\}.& 
a photo of a large \{label\}.& 
a rendition of a \{label\}.&
a photo of a nice \{label\}. \\
a photo of a weird \{label\}.& 
a blurry photo of a \{label\}.&
a cartoon \{label\}.&
art of a \{label\}. \\
a sketch of the \{label\}.& 
a embroidered \{label\}.&
a pixelated photo of a \{label\}.&
itap of the \{label\}. \\
a jpeg corrupted photo of the \{label\}.& 
a good photo of a \{label\}.&
a plushie \{label\}.&
a photo of the nice \{label\}. \\
a photo of the small \{label\}.& 
a photo of the weird \{label\}.&
the cartoon \{label\}.&
art of the \{label\}. \\
a drawing of the \{label\}.& 
a photo of the large \{label\}.& 
a black and white photo of a \{label\}.&
the plushie \{label\}. \\
a dark photo of a \{label\}.& 
itap of a \{label\}.& 
graffiti of the \{label\}.& 
a toy \{label\}. \\
itap of my \{label\}.& 
a photo of a cool \{label\}.&
a photo of a small \{label\}.& 
a tattoo of the \{label\}. \\
\bottomrule
\end{tabular}}
\caption{Full list of prompts to evaluate the performance of zero-shot classification on 11 visual recognition datasets.}
\label{tab:prompt}
\end{table*}

\noindent\textbf{SugarCrepe.}
The SugarCrepe~\cite{parcalabescu2021valse} benchmark encompasses three modalities: Replace, Swap, and Add, enhancing the assessment of visual-linguistic compositional understanding. In the Replace modality, a hard negative is generated by substituting a single atomic concept in a positive caption, which then mismatches with the corresponding image. We classify the replacements according to their conceptual types: REPLACE-OBJ, REPLACE-ATT, and REPLACE-REL. The Swap modality produces hard negatives by interchanging two atomic concepts of the same type within a positive caption, maintaining the original content. This includes SWAP-OBJ and SWAP-ATT categories, excluding relation swaps to avoid generating incoherent text. The Add modality creates hard negatives by introducing a new atomic concept into a positive caption, leading to a discrepancy with the image. It includes ADD-OBJ and ADD-ATT, while relation insertions are excluded due to their implausibility.

\subsection{Details of Negative Caption Generation}
Following CE-CLIP~\cite{zhang2024contrasting}, we adopt 2014 train split of COCO~\cite{chen2015microsoft} as our base dataset and employ the LLM-driven negative caption generation pipeline to generate compositionally-enhanced data for fine-tuning. The base dataset has 83k images and 414K captions. Through our augmentation process, we generate five hard negative captions per original caption, resulting in a total of 2.07 million hard negative captions.\\
\noindent\textbf{Prompt Details.}
In Figure~\ref{fig:prompt}, we detail the prompts employed in this study. Notably, the prompts we designed are concise and effective, especially in ensuring that the generated sentences are ``hard negative'' rather than ``hard positive''. To achieve this, we append the instruction ``ensuring that the rewritten sentences have significantly different or opposite meanings from the original sentences'' at the end of the prompt, which greatly enhanced the quality of both the examples and the final rewritten captions.\\
\noindent\textbf{Examples Generated by ChatGPT.} 
In Figure~\ref{fig:rewritten_example}, we present several representative examples generated by ChatGPT. These examples closely adhere to our rules and serve as high-quality exemplars to guide the large-scale generation of negative samples.

\noindent\textbf{Rewritten Captions.} In Figure~\ref{fig:example}, we present examples of negative samples generated by LLaMA3.1-instruct-8B. These examples demonstrate the high quality of captions produced by the LLM-driven negative caption generation pipeline. The captions adhere accurately to our rewriting rules, ensuring that the generated sentences possess meanings that are distinct from or opposite to the original ones. This efficacy can be attributed to the high-quality examples showcased in Figure~\ref{fig:rewritten_example} and the meticulously designed prompts outlined in Figure~\ref{fig:prompt}.

\end{document}